\pdfoutput=1

\documentclass[11pt]{article}

\usepackage{ACL}

\usepackage{times}
\usepackage{bm}
\usepackage{multirow}
\usepackage{xcolor}
\usepackage{colortbl}

\usepackage{makecell}
\usepackage{color}

\usepackage{algorithm}
\usepackage{booktabs}
\usepackage{amssymb}
\usepackage{amsmath}
\usepackage{float}
\usepackage{times}  % DO NOT CHANGE THIS
\usepackage{helvet}  % DO NOT CHANGE THIS
\usepackage{courier}  % DO NOT CHANGE THIS
\usepackage{graphicx} % DO NOT CHANGE THIS
\usepackage{natbib}  % DO NOT CHANGE THIS AND DO NOT ADD ANY OPTIONS TO IT
\usepackage{caption} % DO NOT CHANGE THIS AND DO NOT ADD ANY OPTIONS TO IT

\usepackage{latexsym}

\usepackage[T1]{fontenc}
\usepackage[utf8]{inputenc}

\usepackage{microtype}

\usepackage{inconsolata}
\usepackage{xcolor}

\usepackage{algpseudocode}

\usepackage{amssymb}
\usepackage{amsmath}
\usepackage{booktabs}
\usepackage{multirow}
\usepackage{xcolor}
\usepackage{colortbl}
\usepackage{soul}
\usepackage{arydshln}
\usepackage{makecell}
\usepackage{newfloat}
\usepackage{listings}

\title{AMATA: Adaptive Multi-Agent Trajectory Alignment for Knowledge-Intensive Question Answering}
\author{Taolin Zhang$^{1}$, Dongyang Li$^{2}$, Chen Chen$^{4}$, Qizhou Chen$^{3}$, Jiuheng Wan$^{1}$, Xiaofeng He$^{3}$, \\ \textbf{Chengyu Wang}$^{5}$\thanks{\ \ C. Wang and R. Hong are co-corresponding authors.}, \textbf{Richang Hong}$^{1}$\footnotemark[1]\\
$^1$ School of Computer Science and Information Engineering, Hefei University of Technology \\
$^2$ Shanghai University of Electric Power 
$^3$ East China Normal University \\
$^4$ Guangdong University of Finance and Economics
$^5$ Alibaba Group\\
 {\tt {tlzhang}@hfut.edu.cn, chengyu.wcy@alibaba-inc.com} \\
 }
\begin{document}
\maketitle

\begin{abstract}
Despite substantial advances in large language models (LLMs), generating factually consistent responses for knowledge-intensive question answering remains challenging. These difficulties are primarily due to hallucinations and the limitations of LLMs in bridging long-tail knowledge gaps. To address this, we propose \emph{AMATA}, an Adaptive Multi-Agent Trajectory Alignment framework that dynamically integrates external knowledge to improve response interpretability and factual grounding. Our architecture leverages six specialized agents that collaboratively perform structured actions for complex question reasoning.
We formalize multi-agent collaboration with external tools as a trajectory preference alignment problem, incorporating question-aware agent customization and inter-agent preference harmonization. AMATA introduces two principal innovations: (1) \emph{Intra-Trajectory Preference Learning}, which learns objective-oriented preferences to prioritize critical agents, and (2) \emph{Inter-Agent Dependency Learning}, which captures cross-agent tool dependencies through a novel dependency-aware direct preference optimization technique. Empirical results show that \emph{AMATA} consistently outperforms baseline approaches, knowledge-augmented frameworks, and LLM-based trajectory systems on five established knowledge-intensive QA benchmarks. Further analysis demonstrates the efficiency of our method in reducing token consumption.
\end{abstract}

\section{Introduction}

Large language models (LLMs) serve as the backbone of modern NLP infrastructure~\cite{DBLP:conf/acl/Hu0Y0ZCC24}, yet they face persistent reliability challenges. Chief among these are hallucinations that appear superficially plausible~\cite{DBLP:conf/naacl/WooZZWGDC25} and other undesirable behaviors~\cite{DBLP:conf/naacl/YangSLSYL25}.

Retrieval-Augmented Generation (RAG) provides a mitigation strategy by enabling LLMs to dynamically retrieve up-to-date information from external knowledge sources during inference~\cite{DBLP:conf/naacl/RubinHB22,DBLP:conf/acl/XuPYMSCZ24}. However, RAG incurs its own drawbacks, such as retrieval inaccuracies~\cite{DBLP:conf/acl/0025FDGY0CCC024} and increased inference latency due to longer contexts~\cite{DBLP:conf/acl/ZouWZJYD24}. As a result, research has increasingly shifted towards multi-agent systems that leverage diverse tooling and cooperative reflection mechanisms to enhance task robustness~\cite{DBLP:conf/acl/XuMYSH24,DBLP:conf/aaai/YueWCHW25}.

\begin{figure*}[!t]
  \centering
  \includegraphics[width=16.25cm]{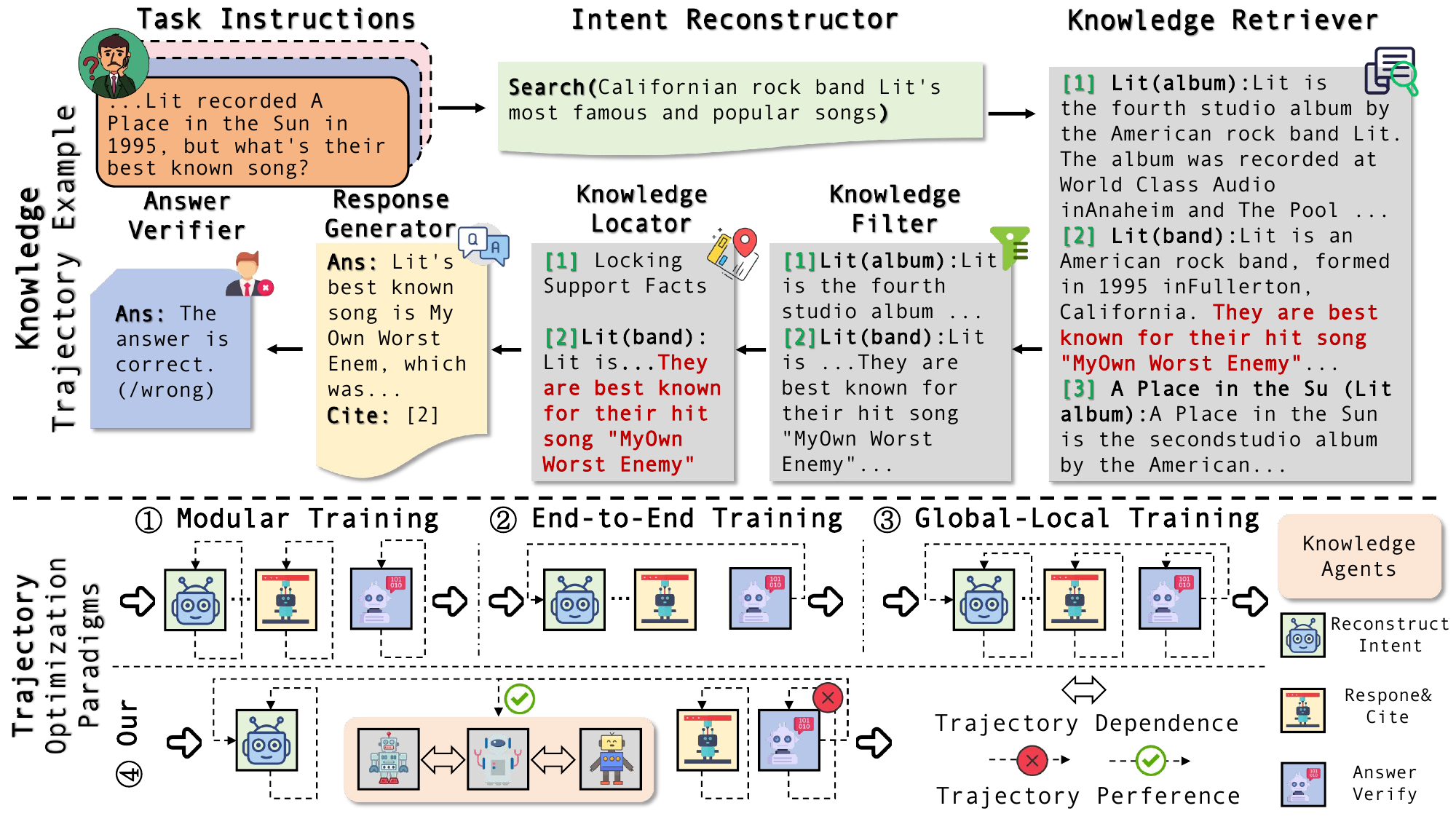}
  \caption{Comparison of multi-agent paradigms for knowledge-intensive QA.  
  \emph{Modular Training} and \emph{End-to-End Training} focus, respectively, on local agent reasoning and global co-adaptation, while the \emph{Global-Local Training} paradigm combines the advantages of both.  
  Our \emph{AMATA} framework dynamically learns intra-trajectory relationships between questions and LLM agents, and establishes inter-agent dependencies across the agent ensemble.}
  \label{motivation_res}
\end{figure*}

Prior multi-agent approaches for knowledge-intensive QA can be broadly categorized into three paradigms: \emph{Modular Training}, \emph{End-to-End Training}, and \emph{Global-Local Training}.  
(1) \emph{Modular Training} fine-tunes individual agents on customized datasets targeted to their respective subtask capabilities~\cite{DBLP:conf/aaai/LongHYLHYL023,DBLP:journals/jodl/KoopmanMLVZGDLZ24}. These locally optimized agents are then manually integrated through static workflows, with execution sequences and handoff mechanisms rigidly defined. The lack of global optimization leads to error propagation and diminished overall performance, as locally optimal agents cannot compensate for inter-agent dependencies or system-wide dynamics.  
(2) \emph{End-to-End Training} adopts a unified optimization strategy, jointly training all agents within a single framework using task-level supervision~\cite{DBLP:conf/emnlp/ZongY0SHCZ24,DBLP:journals/corr/abs-2506-02998}. While backpropagation updates parameters throughout the agent ensemble, enabling co-adaptation and implicit coordination, uniform gradient updates obscure the distinct specialization demands of heterogeneous agents. Agents responsible for different subtasks~\cite{DBLP:conf/iclr/KhotTFF0CS23,DBLP:conf/acl/YueZSLZ024} require individualized learning signals; parameter sharing can cause over-homogenization, thereby eroding specialized expertise.  
(3) \emph{Global-Local Training} employs a two-stage process: first, agents are optimized independently for subtask proficiency, then fine-tuned jointly to align global behaviors~\cite{DBLP:conf/aaai/YueWCHW25}. Although this strategy combines localized specialization with global coordination, it often fails to capture dynamic inter-agent dependencies, which are crucial for handling diverse knowledge-intensive tasks~\cite{DBLP:conf/iclr/ZhangYLYWWCY025,DBLP:journals/corr/abs-2505-00212}. As shown in Figure~\ref{motivation_res}, for questions with high confidence, adding a ``Verifier'' agent after the preceding five agents may be unnecessary. Additionally, the three knowledge agents (``Retriever'', ``Filter'', and ``Locator'') exhibit strong interdependence; when the ``Retriever'' is triggered, subsequent actions of the other two agents must also be executed.

In this paper, we propose \textbf{Adaptive Multi-Agent Trajectory Alignment} (\emph{AMATA}), a framework designed to improve agent-level alignment and capture dynamic inter-agent dependencies. AMATA maintains high reasoning performance while significantly reducing token overhead during inference. Our main contributions are summarized below:

\noindent\textbf{Intra-Trajectory Preference Learning.}  
Existing approaches commonly treat all agents as uniformly relevant throughout the reasoning process. In contrast, our method dynamically optimizes agent participation for each question, learning question-specific agent preference distributions that adaptively modulate each agent's influence based on utility for the current input. For each agent and question pair, we assign a preference score (e.g., \emph{$<$Reconstructor: 5$>$} versus \emph{$<$Verifier: 1$>$}), concatenated with the agent description as a prefix and paired with the question for fine-tuning the corresponding agent.

\noindent\textbf{Inter-Agent Dependency Learning.}  
Multi-agent systems exhibit context-dependent inter-agent dependencies, where triggering a pivotal agent necessitates coordinated execution of functionally linked agents, while allowing conditional suppression of unrelated agents~\cite{DBLP:journals/corr/abs-2409-00335,DBLP:conf/naacl/GaoZCL25}. We introduce a dependency-aware Direct Preference Optimization (DA-DPO) module that learns context-sensitive execution ranking. Specifically, for each question, we construct preference samples that explicitly encode inter-agent dependencies and annotate each trajectory with a joint preference score reflecting the global optimality of the multi-agent sequence. These scores induce a dependency-aware ranking over sampled trajectories, prioritizing those with robust inter-agent coordination for DA-DPO training. This mechanism enables LLMs to infer optimal multi-agent execution sequences with high reliability.

We evaluate \emph{AMATA} against competitive baselines on five benchmarks: HealthQA~\cite{pubhealthtab}, ARC-Choice~\cite{DBLP:journals/corr/abs-1803-05457}, PopQA~\cite{DBLP:journals/corr/abs-2212-10511}, SQuAD 1.1~\cite{DBLP:conf/emnlp/RajpurkarZLL16}, and ASQA~\cite{DBLP:conf/emnlp/GaoYYC23}.  
Our framework achieves an average performance improvement of \textbf{+4.02\%} across all tasks, and reduces token consumption overhead by approximately \textbf{70\%} compared to strong baselines.

\section{Related Work}

\noindent\textbf{Multi-Agent Trajectory Learning.}  
Multi-agent trajectory learning refers to the process of orchestrating multiple agents to collectively solve complex tasks~\cite{DBLP:conf/coling/Li25}. Existing literature can be grouped into three main paradigms:  
(1) \emph{Modular Training}~\cite{DBLP:conf/aaai/LongHYLHYL023,DBLP:journals/jodl/KoopmanMLVZGDLZ24} trains agents independently for their respective subtasks. This often results in suboptimal global performance due to a lack of system-wide coordination. Preference learning has been introduced to ameliorate poor decision-making at the individual agent level~\cite{DBLP:journals/corr/abs-2403-02502,DBLP:conf/emnlp/XiongSZWWWLPL24}, but overall integration remains a challenge.  
(2) \emph{End-to-End Training}~\cite{DBLP:conf/emnlp/ZongY0SHCZ24,DBLP:journals/corr/abs-2506-02998} jointly optimizes all agents using unified loss functions derived from expert trajectories curated by a teacher LLM (e.g., FireAct~\cite{DBLP:journals/corr/abs-2310-05915}, AgentTuning~\cite{DBLP:conf/acl/ZengLLWLD024}). Other approaches such as MapGPT~\cite{DBLP:conf/acl/ChenLXCLW24} and LLM-A$^*$~\cite{DBLP:conf/emnlp/Meng0YPC24} focus on providing agents with a global view of the environment. Despite improved coordination, these methods may obscure contributions from specialized agents, potentially hindering the balance between individual expertise and system-wide collaboration~\cite{DBLP:journals/corr/abs-2505-00212}. 
(3) \emph{Global-Local Training} combines both global context and local adaptation signals to enhance agent specialization. For instance, CoAct~\cite{DBLP:journals/corr/abs-2406-13381} emulates hierarchical human planning in LLMs, while SMART~\cite{DBLP:conf/aaai/YueWCHW25} leverages multi-granular trajectories for agent control and system synergy. These frameworks inject both global and local signals into agent optimization, aiming to preserve both broad task alignment and agent-level differentiation~\cite{DBLP:conf/acl/SubramonianYDB23}. However, these methods often overlook inter-agent dependencies.

\noindent\textbf{Knowledge Enhancement for LLMs.}  
LLMs are prone to hallucinations and lack coverage of long-tail knowledge due to their parametric nature~\cite{DBLP:conf/emnlp/JiYXLIF23,DBLP:conf/acl/LiYZWHHXH24,DBLP:journals/tois/HuangYMZFWCPFQL25}. To ensure factual accuracy, LLMs frequently rely on external sources. RAG incorporates non-parametric resources to improve factual reliability and enrich LLM outputs~\cite{DBLP:conf/kdd/FanDNWLYCL24,DBLP:journals/corr/abs-2501-09136}. Advancements in this area include better retrieval mechanisms using dense retrievers~\cite{DBLP:conf/emnlp/KarpukhinOMLWEC20,DBLP:conf/emnlp/YeLZC24} and improved information integration~\cite{DBLP:conf/ecai/ZhangLC0H0H024,DBLP:journals/corr/abs-2505-11995,DBLP:journals/corr/abs-2503-10677}. For example, Self-RAG~\cite{DBLP:conf/iclr/AsaiWWSH24} introduces reflection tokens to assess both retrieval and response quality during inference. However, these RAG techniques typically operate within a single-agent paradigm, executing retrieval and generation in a sequential pipeline~\cite{DBLP:journals/corr/abs-2501-09136}, thus failing to exploit the emergent capabilities and cooperative reasoning potential of multi-agent LLM frameworks~\cite{qian2024scaling}. In contrast, \emph{AMATA} is designed for knowledge-intensive QA tasks in a multi-agent setting.

\section{Methodology}

\begin{table}[!tb]
\centering
\footnotesize
\begin{tabular}{ccc}
\toprule
\textbf{Agent} & \textbf{Head} & \textbf{End}  \\ \midrule
Intent Reconstructor $\mathcal{A}_{\text{IR}}$ & $\langle\text{Reconstructor}\rangle$ & $\langle\text{/eoi}\rangle$  \\ 
Knowledge Retriever $\mathcal{A}_{\text{KR}}$ & $\langle\text{Retriever}\rangle$ & $\langle\text{/eor}\rangle$  \\ 
Knowledge Filter $\mathcal{A}_{\text{KF}}$ & $\langle\text{Filter}\rangle$ & $\langle\text{/eof}\rangle$  \\ 
Knowledge Locator $\mathcal{A}_{\text{KL}}$ & $\langle\text{Locator}\rangle$ & $\langle\text{/eol}\rangle$  \\ 
Response Generator $\mathcal{A}_{\text{RG}}$ & $\langle\text{Generator}\rangle$ & $\langle\text{/eog}\rangle$  \\  
Answer Verifier $\mathcal{A}_{\text{AV}}$ & $\langle\text{Verifier}\rangle$ & $\langle\text{/eov}\rangle$ \\
\bottomrule
\end{tabular}
\caption{Agents and special tokens used in trajectories. Detailed agent descriptions and the trajectory data collection process are provided in Appendices~\ref{data_metrics} and~\ref{agent_desc_sec}.}
\label{special_token}
\end{table}

\begin{figure*}[!t]
  \centering
  \includegraphics[width=15.75cm]{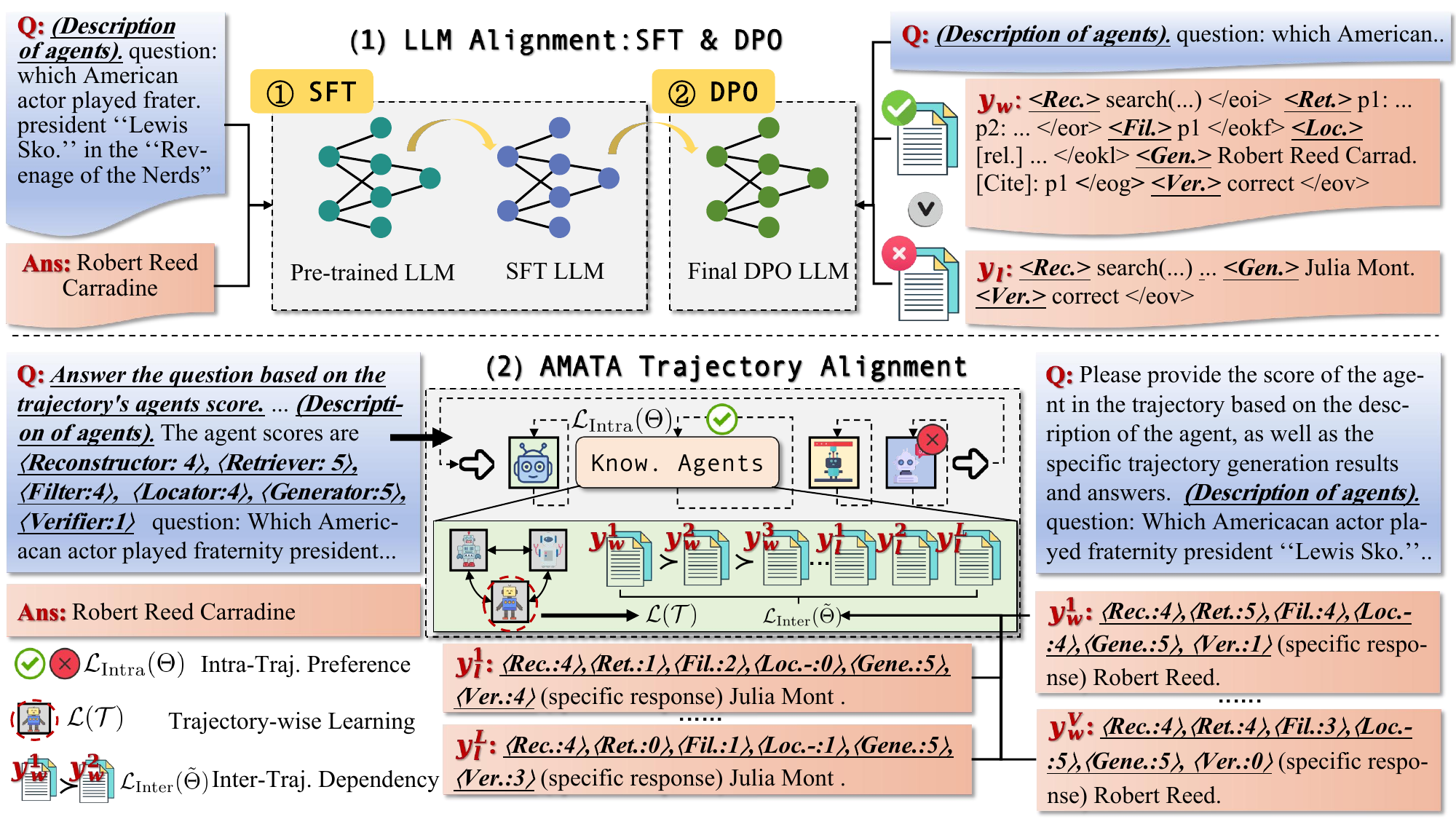}
  \caption{
    Comparison between \emph{AMATA} and standard SFT and DPO pipelines. 
    \emph{AMATA} optimizes intra-trajectory preferences and inter-agent dependencies through adaptive prefix scoring (left) and DA-DPO (right).
  }
  \label{model_overview}
\end{figure*}

\subsection{Task Formulation and Basic Notations}

Figure~\ref{model_overview} illustrates the overall architecture of \emph{AMATA}.  
Given a question $\mathcal{Q}$, we design a workflow utilizing an LLM-based multi-agent system to generate the answer $\mathcal{Y}$, where $\mathcal{Y} = \mathcal{F}(\mathcal{Q}, \mathbf{A})$. Here, $\mathcal{F}$ denotes the entire system parameterized by learnable weights, and $\mathbf{A}$ is the set of agents, such as the six agents in \emph{AMATA} (see Table~\ref{special_token}). In this framework, each agent $\mathcal{A} \in \mathbf{A}$ receives the current state and produces three outputs: a response $y_i$, a special end token $e_i$, and a special head token $h_{i+1}$ for the subsequent agent, expressed as:
\begin{gather}
    y_i, e_i, h_{i+1} = \mathcal{A}(\mathcal{Q}, y_{i-1}, e_{i-1}, h_i),
\end{gather}
where $\mathcal{T} = \{(h_1, y_1, e_1), \dots, (h_T, y_T, e_T)\}$ denotes a complete trajectory realized by dynamically executing the workflow $\mathcal{F}$. The final output $\mathcal{Y}$ is obtained after completing this trajectory. In LLM-based multi-agent systems, conditional autoregressive language modeling is typically adopted to learn which agent should act and when, utilizing these special tokens to coordinate agent behaviors~\cite{DBLP:journals/ral/KwonPJ24,DBLP:journals/corr/abs-2505-17659,DBLP:conf/aaai/YueWCHW25}. The trajectory-wise objective function is defined as
$\mathcal{L}(\mathcal{T}) = \sum_{i=1}^{T} - \log \Pr \left( t_i \mid t_{<i}, \mathcal{Q} \right)$,
where $t_i = (h_i, y_i, e_i)$ represents the $i$-th tuple in the trajectory $\mathcal{T}$ and $t_{<i}$ comprises all preceding tuples in the trajectory.

\subsection{Intra-Trajectory Preference Learning}  

Agents in a multi-agent system exhibit heterogeneous capabilities, necessitating autonomous tool usage and adaptive coordination.
For example, for simple questions, the workflow may not require external knowledge retrieval or output verification (e.g., $\mathcal{A}_{\text{AV}}$).
In such cases, the workflow only formalizes the question via $\mathcal{A}_{\text{IR}}$ and uses the generator to produce an answer via $\mathcal{A}_{\text{RG}}$.
To model agent-specific tool usage within a trajectory, a common approach is supervised fine-tuning (SFT), which enhances the tool-handling skills of individual agents:
\begin{equation}
    \mathcal{L}_{\mathrm{SFT}}^{(j)} (\Theta) = -\mathbb{E}_{(\mathcal{Q}, \mathcal{Y}) \sim \mathcal{D}_{\text{intra}}^{(j)}} \log \Pr(\mathcal{Y} \mid \mathcal{Q}; \Theta),
\end{equation}
where $\mathcal{D}_{\text{intra}}^{(j)}$ is the subset of the intra-trajectory dataset $\mathcal{D}_{\text{intra}}$ corresponding to the $j$-th agent, and $\Theta$ denotes intra-trajectory model parameters.
The agent's description and functionalities are incorporated into $\mathcal{Q}$ via prompt engineering.

While agent-specific training can be effective, it requires both agent-specific datasets and significant computational resources, limiting scalability.
To address this, we consolidate preference learning within a unified trajectory sampling framework, enabling a monolithic model to capture heterogeneous agent competencies through SFT augmented with adaptive prefix scoring:
\begin{gather}
    \mathcal{L}_{\mathrm{Intra}}(\Theta) = - \mathbb{E}_{(\mathcal{Q}, \mathcal{Y}, P) \sim \mathcal{D}_{\text{intra}}} \log \Pr (\mathcal{Y} \mid \mathcal{P}, \mathcal{Q}; \Theta)
\end{gather}
where $\mathcal{P} = \{P_{\mathcal{A}_{\text{IR}}}, \dots, P_{\mathcal{A}_{\text{AV}}}\}$ represents the preference prefix set for each agent in the trajectory, indicating their relative importance for a given sample (e.g., \emph{$<$Retriever: 5$>$}).
These scores reflect the importance of each agent in correctly answering the question, as annotated by LLMs.\footnote{Prompt templates are provided in Appendix~\ref{data_metrics}.} 

\begin{figure*}[!t]
  \centering
  \includegraphics[width=16.25cm]{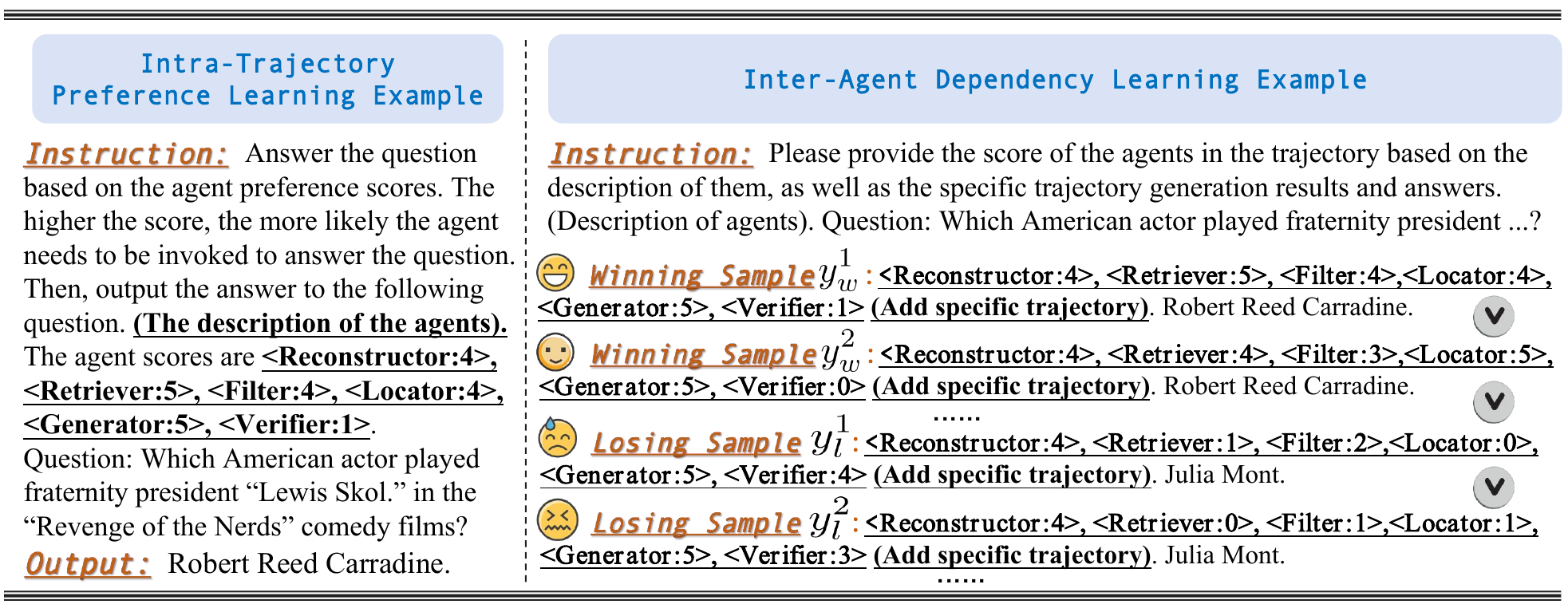}
  \caption{Two-stage training examples of \emph{AMATA}. Detailed annotation process and robustness verification for the \textbf{\emph{DEPENDENCY SCORES}} are provided in Appendix~\ref{trajectory_scores_app}.}
  \label{instruction_example}
\end{figure*}

Consider the intra-trajectory sample illustrated in Fig.~\ref{instruction_example}.
The reference to ``Revenge of the Nerds'' requires background knowledge to support the LLM's response. Accordingly, the \texttt{$\langle$Retriever$\rangle$}, \texttt{$\langle$Filter$\rangle$}, and \texttt{$\langle$Locator$\rangle$} agents receive higher preference scores, reflecting a stronger need for knowledge tools.
When knowledge agents yield high-confidence outputs, verification becomes redundant, resulting in a low preference score for the \texttt{$\langle$Verifier$\rangle$} agent.
This example demonstrates that question complexity induces a dynamic, heterogeneous agent hierarchy within the reasoning trajectory, and underscores the need for a framework that supports flexible, token-efficient tool orchestration.

\subsection{Inter-Agent Dependency Learning}

Functional dependencies naturally exist between agents within a trajectory, as downstream agents rely on responses and states generated by upstream agents in order to be triggered and operate effectively~\cite{DBLP:conf/naacl/GaoZCL25}.
In \emph{AMATA}, such dependencies are especially pronounced: for instance, if the retrieval agent $\mathcal{A}_{\text{KR}}$ is activated, then the knowledge filtering agent $\mathcal{A}_{\text{KF}}$ and the locating agent $\mathcal{A}_{\text{KL}}$ must also be executed.
To model these interactions, we introduce inter-agent dependency learning, leveraging pairs of winning and losing samples for Direct Preference Optimization (DPO)~\cite{DBLP:conf/nips/RafailovSMMEF23}.
This enables the model to automatically discover optimal patterns of agent collaboration and to capture the underlying dependency structures among agents.

A widely adopted technique combines the Bradley--Terry (BT) model~\cite{BT_model} with DPO to parameterize the reward function for trajectory selection.
Formally, the probability that the winning response $y_w$ is preferred over the losing response $y_l$ for a given instruction $\mathcal{Q}$ is:
\begin{equation}
\Pr(y_w \succ y_l \mid \mathcal{Q}) = \frac{\exp(r(\mathcal{Q}, y_w))}{\exp(r(\mathcal{Q}, y_w)) + \exp(r(\mathcal{Q}, y_l))}
\end{equation}
where the reward $r(\mathcal{Q}, y)$ measures the policy model's preference for $y$ and is given by
$r(\mathcal{Q}, y) = \beta \cdot \log \frac{\pi_{\tilde{\Theta}}(y \mid \mathcal{Q})}{\pi_{\mathrm{ref}}(y \mid \mathcal{Q})} + \beta \cdot \log \mathrm{Z}(\mathcal{Q})$,
with $\pi_{\mathrm{ref}}$ and $\pi_{\tilde{\Theta}}$ denoting the intra-trajectory and inter-agent dependency models, respectively.
The coefficient $\beta$ modulates the strength of regularization, and $\mathrm{Z}(\mathcal{Q}) = \sum_{y} \pi_{\mathrm{ref}}(y \mid \mathcal{Q}) \exp\left(\frac{1}{\beta} r(\mathcal{Q}, y)\right)$ denotes the partition function.

However, merely distinguishing between winning and losing samples is insufficient for optimal dependency modeling.
Our analysis (see Sect.~\ref{detailed_analysis}) reveals that specific combinations of agent preference scores in the trajectory prefix are correlated with higher response quality, compared to settings that neglect agent dependency information.
For example, in the inter-trajectory samples presented in Fig.~\ref{instruction_example}, although both $y_w^1$ and $y_w^2$ yield correct responses, $y_w^1$ should receive a higher trajectory policy preference due to its consistently elevated scores for tightly coupled agents (such as \texttt{$\langle$Retriever$\rangle$}, \texttt{$\langle$Filter$\rangle$}, and \texttt{$\langle$Locator$\rangle$}).
Moreover, both of these trajectories outperform all losing samples.
To address this, we propose the \emph{dependency-aware DPO} (DA-DPO) algorithm in \emph{AMATA}, guiding the policy model to better capture relative dependency relationships among agent preference scores.

\begin{table*}[!tb]
\centering
\footnotesize
\setlength{\tabcolsep}{3pt}
    \begin{tabular}{lccccccccc}
      \toprule
      \multicolumn{1}{c}{\textbf{Task} $\rightarrow$}  & \textbf{HealthQA} & \textbf{ARC-C} & \textbf{PopQA} & \textbf{Squad1} & \multicolumn{3}{c}{\textbf{ASQA}} & \multirow{2}{*}{\textbf{Average}} \\  
       \multicolumn{1}{c}{\textbf{Model} $\downarrow$}  & Acc. & Acc. & Acc. & Acc. & Str\_EM & Rouge-L & Mauve & \\  \midrule
      \rowcolor{lightgray} \multicolumn{9}{c}{\texttt{\textbf{Vanilla QA Methods}}} \\ \midrule
\multicolumn{1}{c}{ $\text{Alpaca2}\text{ 7B}$} & 44.78$_{(\pm1.2)}$ & 36.43$_{(\pm1.5)}$ & 25.58$_{(\pm0.8)}$ & 11.50$_{(\pm1.1)}$ & 14.42$_{(\pm1.6)}$ & 28.72$_{(\pm2.1)}$ & 51.24$_{(\pm0.9)}$ & 30.38$_{(\pm1.1)}$ \\
\multicolumn{1}{c}{ $\text{Mistral-Instruct}\text{ 7B}$} & 65.45$_{(\pm1.4)}$ & 57.84$_{(\pm0.7)}$ & 22.37$_{(\pm1.3)}$ & 14.97$_{(\pm1.3)}$ & 20.80$_{(\pm2.2)}$ & 32.20$_{(\pm0.9)}$ & 33.47$_{(\pm1.8)}$ & 35.30$_{(\pm0.8)}$ \\
\multicolumn{1}{c}{ $\text{Llama-2-Chat}\text{ 7B}$ } & 47.95$_{(\pm1.9)}$ & 47.95$_{(\pm1.1)}$ & 25.44$_{(\pm0.8)}$ & 14.13$_{(\pm0.7)}$ & 16.79$_{(\pm2.3)}$ & 32.35$_{(\pm1.6)}$ & 24.21$_{(\pm1.2)}$ & 29.83$_{(\pm1.5)}$ \\
\multicolumn{1}{c}{ $\text{Vicuna-v1.5}\text{ 13B}$ } & 63.01$_{(\pm2.0)}$ & 57.59$_{(\pm0.9)}$ & 17.94$_{(\pm1.5)}$ & 15.25$_{(\pm1.8)}$ & 31.95$_{(\pm2.2)}$ & 22.99$_{(\pm1.7)}$ & 68.41$_{(\pm1.4)}$ & 39.59$_{(\pm1.3)}$ \\
\multicolumn{1}{c}{ $\text{Llama-2-Chat}\text{ 13B}$} & 62.20$_{(\pm1.8)}$ & 48.72$_{(\pm2.2)}$ & 21.22$_{(\pm1.9)}$ & 15.97$_{(\pm1.4)}$ & 19.97$_{(\pm0.7)}$ & 30.37$_{(\pm1.3)}$ & 40.23$_{(\pm1.5)}$ & 34.10$_{(\pm1.7)}$ \\
\multicolumn{1}{c}{ $\text{Qwen-2.5-Ins.}\text{ 7B}$} & 64.02$_{(\pm1.1)}$ & 51.38$_{(\pm1.7)}$ & 22.35$_{(\pm0.7)}$ & 17.23$_{(\pm1.2)}$ & 18.99$_{(\pm1.3)}$ & 31.65$_{(\pm2.0)}$ & 47.03$_{(\pm1.1)}$ & 36.09$_{(\pm1.2)}$ \\
\multicolumn{1}{c}{\it $\text{GPT-3.5-turbo}$} & \it 76.08$_{(\pm1.5)}$ & \it 77.30$_{(\pm0.8)}$ & \it 29.30$_{(\pm1.2)}$ & \it 22.90$_{(\pm1.6)}$ & \it 39.94$_{(\pm1.4)}$ & \it 35.73$_{(\pm0.7)}$ & \it 44.63$_{(\pm2.3)}$ & \it 46.55$_{(\pm1.4)}$ \\
\midrule
\rowcolor{lightgray} \multicolumn{9}{c}{\texttt{\textbf{Knowledge-augmented Methods}}} \\ \midrule
\multicolumn{1}{c}{$\text{Alpaca2}\text{ 7B}$} & 26.44$_{(\pm1.7)}$ & 35.15$_{(\pm1.4)}$ & 33.38$_{(\pm1.6)}$ & 21.41$_{(\pm2.2)}$ & 23.59$_{(\pm0.8)}$ & 27.21$_{(\pm2.3)}$ & 50.09$_{(\pm1.5)}$ & 31.04$_{(\pm1.2)}$ \\
\multicolumn{1}{c}{$\text{REPLUG}_\text{ 7B}$} & 41.72$_{(\pm1.3)}$ & 47.26$_{(\pm0.8)}$ & 37.24$_{(\pm1.5)}$ & 24.23$_{(\pm1.7)}$ & 26.54$_{(\pm2.2)}$ & 33.25$_{(\pm0.9)}$ & 54.03$_{(\pm1.4)}$ & 37.75$_{(\pm1.7)}$ \\
\multicolumn{1}{c}{$\text{VANILLA}\text{ 7B*}$} & 29.52$_{(\pm1.6)}$ & 42.74$_{(\pm2.2)}$ & 37.52$_{(\pm1.7)}$ & 25.92$_{(\pm1.4)}$ & 32.25$_{(\pm1.5)}$ & 34.93$_{(\pm0.8)}$ & 39.54$_{(\pm2.3)}$ & 34.63$_{(\pm1.1)}$ \\
\multicolumn{1}{c}{$\text{RADIT}\text{ 7B}$} & 52.98$_{(\pm1.5)}$ & 62.10$_{(\pm1.7)}$ & 38.02$_{(\pm2.3)}$ & 23.86$_{(\pm0.8)}$ & 25.68$_{(\pm1.4)}$ & 15.99$_{(\pm1.6)}$ & 12.35$_{(\pm1.9)}$ & 33.00$_{(\pm1.2)}$ \\
\multicolumn{1}{c}{$\text{INTERACT}_\text{ 7B}$} & 65.45$_{(\pm0.8)}$ & 48.12$_{(\pm1.3)}$ & 41.31$_{(\pm1.6)}$ & \underline{31.52}$_{(\pm1.4)}$ & 34.54$_{(\pm1.7)}$ & 35.51$_{(\pm2.2)}$ & 43.45$_{(\pm1.5)}$ & 42.84$_{(\pm0.9)}$ \\
\multicolumn{1}{c}{$\text{SelfRag}\text{ 7B*}$} & 68.99$_{(\pm1.4)}$ & 65.52$_{(\pm0.6)}$ & 40.67$_{(\pm0.8)}$ & 22.39$_{(\pm1.3)}$ & 28.68$_{(\pm1.5)}$ & 34.11$_{(\pm1.7)}$ & 83.00$_{(\pm2.1)}$ & 49.05$_{(\pm1.8)}$ \\ \midrule
\rowcolor{lightgray} \multicolumn{9}{c}{\texttt{\textbf{LLM-based Trajectory Methods}}} \\ \midrule
\multicolumn{1}{c}{$\text{MMAgent}\text{ 7B}$ } & 72.56$_{(\pm0.7)}$ & 64.43$_{(\pm1.3)}$ & 37.92$_{(\pm1.5)}$ & 24.62$_{(\pm0.8)}$ & 34.13$_{(\pm1.6)}$ & 37.25$_{(\pm1.2)}$ & 90.11$_{(\pm2.0)}$ & 51.57$_{(\pm1.3)}$ \\
\multicolumn{1}{c}{ $\text{SMART}\text{ 7B}$ } & 73.90$_{(\pm1.6)}$ & 67.31$_{(\pm1.4)}$ & 42.88$_{(\pm0.8)}$ & 29.24$_{(\pm1.3)}$ & 42.56$_{(\pm1.7)}$ & 41.71$_{(\pm1.5)}$ & 92.32$_{(\pm2.2)}$ & 55.70$_{(\pm2.1)}$ \\
\multicolumn{1}{c}{ $\text{SPA-RL}\text{ 7B}^{\dagger}$} & {73.23}$_{(\pm1.2)}$ & \underline{68.53}$_{(\pm0.7)}$ & {42.72}$_{(\pm2.1)}$ & 29.46$_{(\pm1.6)}$ &{43.73}$_{(\pm0.8)}$ & {41.37}$_{(\pm1.3)}$ & {91.02}$_{(\pm1.5)}$& 55.72$_{(\pm2.6)}$ \\
\multicolumn{1}{c}{ $\text{GiGPO}\text{ 7B}^{\dagger}$} & \underline{73.97}$_{(\pm1.5)}$ & 68.01$_{(\pm1.8)}$ & \underline{43.52}$_{(\pm1.3)}$ & {29.97}$_{(\pm1.7)}$ & \underline{43.88}$_{(\pm1.4)}$ & \underline{43.64}$_{(\pm1.1)}$ & \underline{92.83}$_{(\pm1.2)}$ & \underline{56.55}$_{(\pm2.1)}$ \\
\multicolumn{1}{c}{$\text{AMATA}\text{ 7B}$} & \textbf{75.83}$_{(\pm0.8)}$ & \textbf{72.47}$_{(\pm1.6)}$ & \textbf{47.39}$_{(\pm1.2)}$ & \textbf{34.61}$_{(\pm1.5)}$ & \textbf{49.10}$_{(\pm1.7)}$ & \textbf{48.26}$_{(\pm1.3)}$ & \textbf{96.35}$_{(\pm1.4)}$ & \textbf{60.57$_{(\pm1.1)}$} \\
\bottomrule
\end{tabular}
\caption{Overall results of \emph{AMATA}. Results of GPT-3.5-turbo are for reference only. $*$ indicates re-implemented methods based on the same model. $\dagger$ denotes the RL settings described in Appendix~\ref{training_details}. Results for other LLMs are shown in Appendix~\ref{backbone_result}. \textbf{Bold} numbers represent the best results, while \underline{underlined} indicate the second-best.}
\label{main_res}
\end{table*}

Specifically, given $M$ winning and $N$ losing samples for a question $\mathcal{Q}$, we first select the top-$K$ winning samples based on their \emph{dependency scores}, which measure the ``goodness'' of dependency among agent preference scores as determined by prior knowledge.  
We treat the remaining $M-K$ winning samples as losing ones due to their lower ``goodness'' of dependencies. These samples are denoted as $(y_w^1, \dots, y_w^{K}, y_w^{K+1}, \dots, y_w^{M})$ and $(y_l^{M+1}, \dots, y_l^{M+N})$, where $(y_w^{K+1}, \dots, y_w^{M})$ and $(y_l^{M+1}, \dots, y_l^{M+N})$ are treated as losing samples. Inspired by listwise Plackett-Luce preference modeling~\cite{list-wise-PL}, we define the inter-agent dependency model as follows:
\begin{equation}
\small
\begin{aligned}
&\Pr\left(y_w^1 \succ y_w^2 \succ \cdots \succ y_w^{K} \succ \{y_w^{K+1}, \dots, y_l^{M+N}\} \mid \mathcal{Q} \right) \\
&= \sum_{f_{K+1}^{M+N}} \prod_{i=1}^{M+N-1} f_\mathcal{E}(\mathcal{Q}, y_i)\\
&= \prod_{i=1}^{K} f_\mathcal{E}(\mathcal{Q}, y_i) 
   \cdot \sum_{f_{K+1}^{M+N}} \prod_{i=K+1}^{M+N-1} f_\mathcal{E}(\mathcal{Q}, y_i) \\
&= \prod_{i=1}^{K} f_\mathcal{E}(\mathcal{Q}, y_i) 
   \cdot \sum_{f_{K+1}^{M+N}} \Pr(y_{K+1} \succ \cdots \succ y_{M+N} \mid \mathcal{Q}) \\
&= \prod_{i=1}^{K} f_\mathcal{E}(\mathcal{Q}, y_i)
\end{aligned}
\end{equation}
where $f_{K+1}^{M+N}$ denotes the set of all permutations of $(y_{K+1}, \ldots, y_{M+N})$ and $f_\mathcal{E}(\mathcal{Q}, y_i) = \frac{exp(r(\mathcal{Q}, y))}{\sum_{j=i}^{M+N} exp(r(\mathcal{Q}, y))}$.
The set $\{y_w^{K+1}, \dots, y_l^{M+N}\}$ denotes the rejected trajectory set for inter-agent dependency learning, including $(M-K)$ original winning samples and $N$ losing samples.  

By substituting the reward from Eq.~(7) into the probability maximization objective 
$\Pr(y_w^1 \succ \cdots \succ y_w^{K} \succ \{y_w^{K+1}, \dots, y_l^{M+N}\} \mid \mathcal{Q})$, we obtain the objective of the inter-agent dependency loss:
\begin{gather}
\mathcal{L}_\text{Inter}(\tilde{\Theta})
= - \mathbb{E}_{(\mathcal{Q}, y_w^1, \ldots, y_l^L) \sim \mathcal{D}^{\text{inter}}} \mathcal{H}(\mathcal{Q}, y) \\
\mathcal{H}(\mathcal{Q}, y) = \sum_{i=1}^{K} \log \sigma \left( - \log \sum_{j=i+1}^{M+N} \exp \mathcal{V}_\beta \right)
\end{gather}
where $\mathcal{V}_\beta = \beta \log \frac{\pi_{\tilde{\Theta}}(y_j \mid \mathcal{Q})}{\pi_{\mathrm{ref}}(y_j \mid \mathcal{Q})} - \beta \log \frac{\pi_{\tilde{\Theta}}(y_i \mid \mathcal{Q})}{\pi_{\mathrm{ref}}(y_i \mid \mathcal{Q})}$,
and $\mathcal{D}^{\text{inter}}$ and $\tilde{\Theta}$ denote the inter-agent training samples and parameters, respectively.

\subsection{Model Training and Inference}
\label{model_training}
Our \emph{AMATA} framework undergoes two-stage training. First, we perform intra-trajectory preference learning using $\mathcal{L}_{\text{Intra}}(\Theta)$, which enables the model to acquire varying degrees of perception regarding agent utilization within trajectories.  
Next, we combine the basic agent prediction loss $\mathcal{L}(\mathcal{T})$ and the inter-agent dependency loss $\mathcal{L}_{\text{Inter}}(\tilde{\Theta})$ to form the total loss:
\begin{equation}
\mathcal{L}_{\text{total}} = \alpha_1 \cdot \mathcal{L}(\mathcal{T}) + \alpha_2 \cdot \mathcal{L}_{\text{Inter}}(\tilde{\Theta}),
\end{equation}
thereby enhancing multi-agent cooperation in knowledge-intensive QA tasks.  
Here, $\alpha_1$ and $\alpha_2$ are training coefficients that sum to 1.  
Due to space constraints, we refer readers to Appendix~\ref{inference_process} for inference algorithm details.

% \footnote{The inference process is shown in Appendix \ref{inference_process}.}

% During inference, \emph{AMATA} first analyzes the query to decide if external knowledge is needed. If not, it directly generates and verifies the answer. Otherwise, it retrieves and filters relevant documents and generates a grounded response. The answer is then verified, and if incorrect, the process iterates with updated instructions to refine the response. Refer to the Appendix \ref{inference_process} for algorithmic details.

\section{Experiments}
We conduct extensive experiments to evaluate \emph{AMATA}. Due to space limitations, details regarding trajectory data collection, baselines, and implementation are provided in Appendix~\ref{experimental_res}.

\subsection{Main Results}
As shown in Table~\ref{main_res}, the key observations are as follows:
(1) Compared to standard QA baselines, \emph{AMATA} significantly outperforms models with comparable or even larger parameter sizes. Notably, it incorporates external knowledge through trajectory learning, compensating for the parameter-size gap relative to larger backbones such as Vicuna-v1.5 (13B) and Llama2-13B-Chat, especially on long-tail knowledge tasks (i.e., PopQA, SQuAD 1.1, and ASQA) in comparison to GPT-3.5-turbo.
(2) Knowledge-augmented methods leverage retrievers to access external data and assist LLMs in generating informed answers; however, data noise and excessive augmentation can significantly degrade model performance~\cite{fangenhancing}. Notably, despite sharing the same training data and backbone as RADIT~\cite{DBLP:conf/iclr/Lin0CSL00KSLZY24}, our model achieves a higher fluency score as measured by \textit{MAUVE}.
(3) We also compare against LLM-based trajectory methods, including the independent-agent modular training method MMAgent (comprising six independent agents), the global-local trajectory approach SMART~\cite{DBLP:conf/aaai/YueWCHW25}, and recent long-trajectory methods such as GiGPO~\cite{DBLP:journals/corr/abs-2505-10978} and SPA-RL~\cite{DBLP:journals/corr/abs-2505-20732}. The results indicate that long-trajectory methods generally outperform multi-agent training approaches in our setting. This improvement can be attributed to RL feedback, which mitigates the cumulative propagation of agent-action errors in longer trajectories. Our method further encourages both intra- and inter-agent dependencies within trajectories, thereby reducing collaborative conflicts among agents.

\begin{table}[!tb]
\centering
\small
    \begin{tabular}{lcccc}
      \toprule
      \multicolumn{1}{c}{\textbf{Task} $\rightarrow$}  & \textbf{HealthQA} & \textbf{ARC-C} & \textbf{PopQA} & \textbf{ASQA} \\
     \multicolumn{1}{c}{ \textbf{Model} $\downarrow$}  & \textbf{Acc.} & \textbf{Acc.} & \textbf{Acc.} & \textbf{Str\_EM}  \\
      \midrule
      \rowcolor{lightgray} \multicolumn{5}{c}{\texttt{\textbf{Training ablation}}} \\ \midrule
   \multicolumn{1}{c}{     $\text{AMATA}_\text{ 7B}$}   & 75.83 & 72.47 & 47.39 & 49.10  \\ \midrule
    \multicolumn{1}{c}{w/o $    \mathcal{L}_{\mathrm{Intra}}$ }   & 72.91 & 70.05 & 44.32 & 45.28  \\
     \multicolumn{1}{c}{ w/o $    \mathcal{L}_{\mathrm{Inter}}$  } & 70.86 & 67.57 & 41.13 & 42.94  \\
     \multicolumn{1}{c}{ w/o $    \mathcal{L}_{\mathcal{T}}$  } & 72.03 & 69.78 & 43.24 & 44.62  \\ \hdashline
     \multicolumn{1}{c}{  w/o $\mathcal{A}_{\text{IR}}$}   & 73.38 & 70.50 & 45.41 & 46.95  \\
          \multicolumn{1}{c}{  w/o $\mathcal{A}_{\text{KF}}$ }      &  72.66 & 69.98 & 44.27 & 45.12   \\ 
\multicolumn{1}{c}{  w/o $\mathcal{A}_{\text{AV}}$ } &  73.13 & 69.34 & 45.20 & 45.85 \\ \midrule
      \rowcolor{lightgray} \multicolumn{5}{c}{\texttt{\textbf{Inference ablation}}} \\ \midrule
  \multicolumn{1}{c}{  w/o $\mathcal{A}_{\text{IR}}$}   & 73.57 & 70.83 & 45.75 & 47.26  \\
 \multicolumn{1}{c}{  w/o $\mathcal{A}_{\text{KF}}$ }      & 72.99 & 70.15 & 44.89 & 45.37  \\ 
\multicolumn{1}{c}{  w/o $\mathcal{A}_{\text{AV}}$ } & 73.36 & 70.29 & 45.41 & 47.50 \\     \bottomrule
    \end{tabular}
\caption{Training and inference ablation of key trajectory learning modules and agents in \emph{AMATA}.}
\label{ablation_study}
\end{table}

\subsection{Ablation Study}
We perform ablation studies on the critical modules involved in the training and inference processes of \emph{AMATA}.  
As shown in Table~\ref{ablation_study}, during training, removing $\mathcal{L}_{\text{Inter}}$ prevents the modeling of associations between agents (e.g., knowledge agents and verifiers), resulting in unresolved collaborative conflicts and the steepest decline in model performance.
Additionally, removing the basic trajectory loss $\mathcal{L}(\mathcal{T})$ causes a significant performance drop, as it impairs the semantic modeling of agent trajectories.

From the perspective of individual agents, removing $\mathcal{A}_{\text{KF}}$ fails to effectively reduce noise in the data retrieved by $\mathcal{A}_{\text{KR}}$, leading to a marked decline in performance.
Removing the verifier agent $\mathcal{A}_{\text{AV}}$ eliminates the verification of generated trajectory answers, which may cause hallucinated outputs and subsequently degrade results.
Due to variability in question content, $\mathcal{A}_{\text{IR}}$ is essential for enabling semantic understanding of user queries; thus, its removal also adversely affects overall performance.

\subsection{Detailed Analysis}
\label{detailed_analysis}

In this section, we conduct an in-depth analysis of the adaptive cooperation among \emph{AMATA} agents, elucidating its advantages in both performance and efficiency. Due to space limitations, \textbf{computational cost comparison}, \textbf{LLM backbones}, \textbf{hyperparameter analysis}, and \textbf{case studies} are presented in Appendix~\ref{additional_results}.

\begin{figure}[!t]
  \centering
  \includegraphics[height=4.5cm, width=8cm]{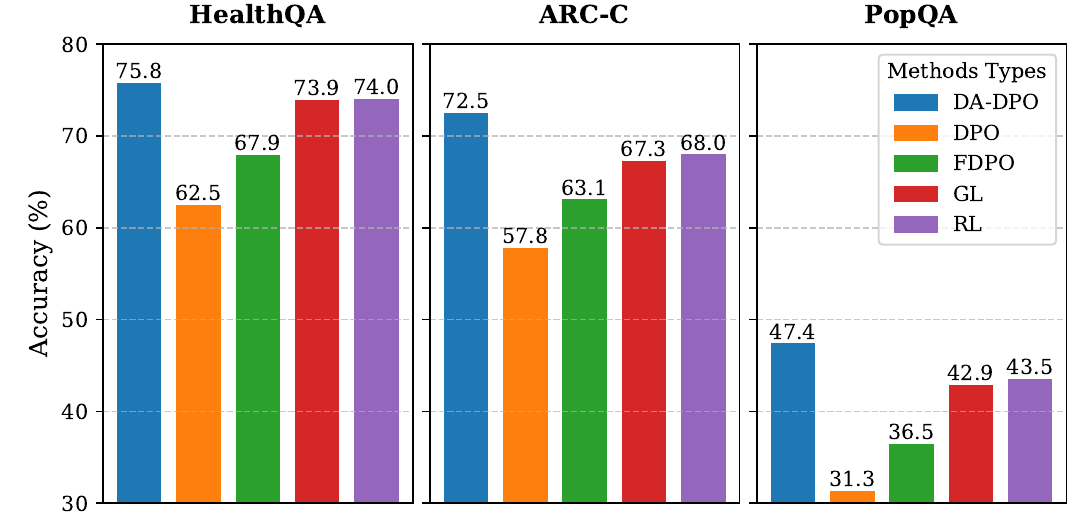}
  \caption{Agent dependency analysis across preference methods. ``DA-DPO'', ``FDPO'', ``GL'', and ``RL'' refer to our DA-DPO, full-order DPO, global-local trajectory, and reinforcement learning, respectively.}
  \label{agents_dependency}
\end{figure}

\noindent\textbf{Agent Dependency Analysis.}  
We adopt different DPO methods to investigate whether comparing winning and losing examples can enhance dependencies among agents.  
Specifically, we compare our DA-DPO method with DPO~\cite{DBLP:conf/nips/RafailovSMMEF23} and full-order DPO (FDPO)~\cite{DBLP:conf/nips/RafailovSMMEF23}.  
The DPO method utilizes only a single pair of winning and losing QA samples, while FDPO leverages the same number of samples as our method but ranks them using full-order learning derived from the magnitude of preference scores.  
Additionally, we compare our DA-DPO with global-local (SMART) and RL-based (GiGPO) approaches.

In Figure~\ref{agents_dependency}, our DA-DPO method consistently outperforms the other DPO-based approaches. This improvement is attributed to its fine-grained preference learning for trajectories in multi-agent collaboration, guided by preference scores.  
Subtle differences in scores for winning examples effectively reflect strong correlations between agents, while the inclusion of losing samples and winning examples with weaker scores helps distinguish less dependent relationships.  
Methods that do not explicitly account for pairwise preference data or that rely on FDPO-style comprehensive sorting tend to reduce overall model effectiveness.  
Regarding the global-local trajectory and RL-based approaches, their results exhibit greater consistency compared to DPO and FDPO. We hypothesize that this improvement arises because global-local fusion and RL step-level feedback enhance supervision of fine-grained agent dependencies~\cite{DBLP:conf/emnlp/DuLWFLW24}.

\begin{figure}[!t]
  \centering
  \includegraphics[width=7.5cm]{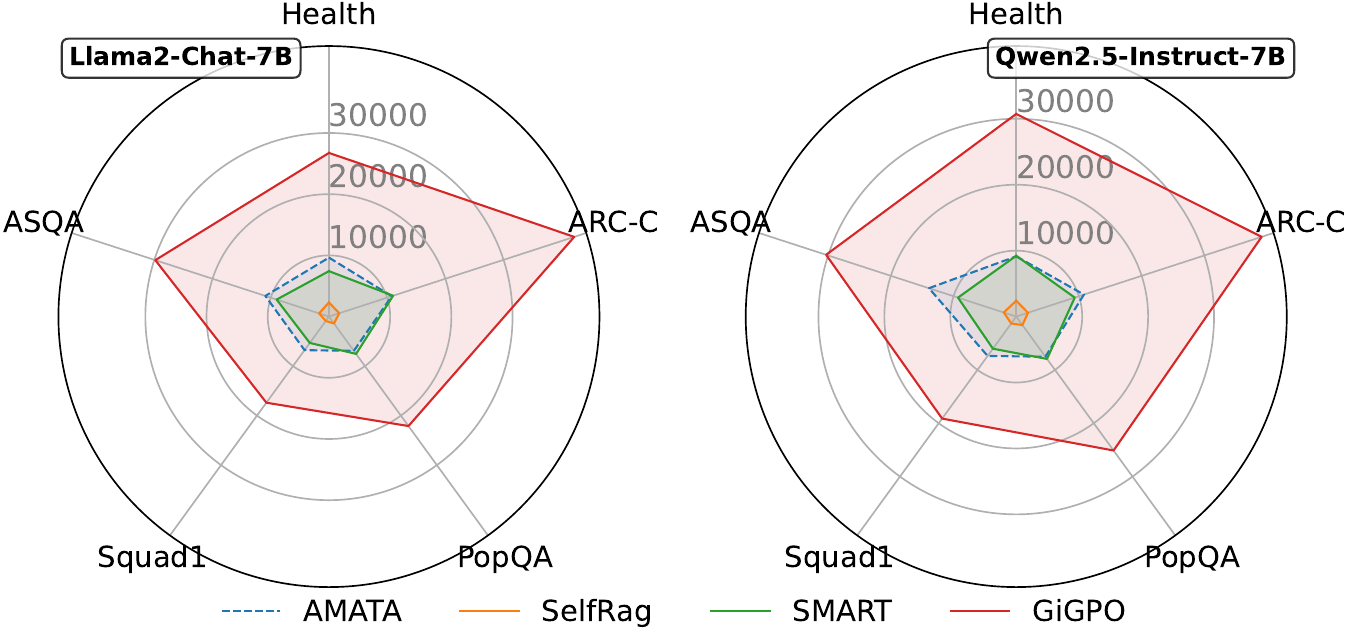}
  \caption{Average token consumption per question.}
  \label{token_number}
\end{figure}

\begin{figure}[!t]
  \centering
  \includegraphics[height=9cm]{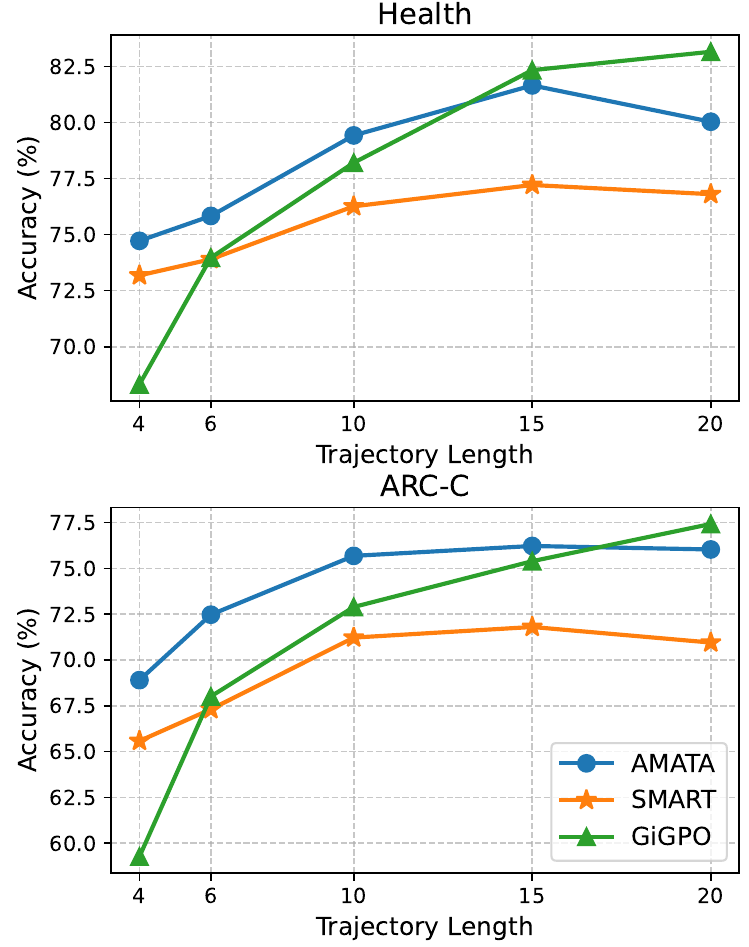}
  \caption{Performance of LLM-based trajectory methods relative to trajectory length.}
  \label{trajectory_length}
\end{figure}

\noindent\textbf{Token Consumption.}  
In Figure~\ref{token_number}, we compare the number of tokens processed by two base models (Llama2-7B and Qwen2.5-7B) when solving questions across different multi-agent frameworks. We observe that: 
(1) Although knowledge-augmented methods consume fewer tokens via a single enhancement step, their performance remains modest, as shown in Table~\ref{main_res}.  
(2) Compared with the global-local trajectory baseline SMART~\cite{DBLP:conf/aaai/YueWCHW25}, although SMART involves fewer agent interactions, \emph{AMATA} not only maintains comparable token overhead but also significantly outperforms it in terms of performance (+4.87\%).  
This improvement stems from our adaptive agent preference learning, which enables \emph{AMATA} to dynamically adjust interactions among knowledge agents (i.e., $\mathcal{A}_{\text{KR}}, \mathcal{A}_{\text{KF}}$, and $\mathcal{A}_{\text{KL}}$) and other agents, greatly reducing token consumption.  
(3) By contrast, RL-based methods require multiple rollout sessions to compute advantages, resulting in the highest token consumption (+70\%).  
Moreover, due to the absence of real-world step-level reward signals, these methods introduce noisy data to the knowledge agents, leading to reduced performance (-4.02\%).

\noindent\textbf{Generalization by Trajectory Length.}  
Figure~\ref{trajectory_length} compares the performance of LLM-based trajectory methods as trajectory length increases, i.e., as the number of agents in the multi-agent system grows.  
Constructing longer trajectories simulates multi-agent environments by introducing more agents to evaluate whether retrieved documents are properly filtered and located.  
We observe that when the number of agents is relatively small, our adaptive preference learning method and the global-local trajectory method outperform the RL approach.  
We hypothesize that shorter trajectories facilitate simpler and more direct end-to-end supervised training.  
By contrast, RL-based methods require the computation of advantage values based on feedback at each step, which can hinder the timely propagation of feedback to local agents.  
However, as trajectory length increases, the RL-based method gradually improves results due to the cumulative effect of reward feedback and iterative rollout learning.  
Meanwhile, other methods suffer from excessive error accumulation, resulting in performance degradation.  
In this paper, knowledge QA tasks generally involve short trajectories (fewer than 10 steps), while GUI tasks typically involve long trajectories (more than 50 steps)~\cite{DBLP:conf/nips/Yao0YN22}.

\section{Conclusion}
We propose an adaptive multi-agent trajectory framework, \emph{AMATA}, which enhances LLMs by effectively incorporating external knowledge to solve knowledge-intensive QA tasks.  
Our key innovations in intra-trajectory and inter-agent preference learning enable prioritization of critical agents and accurate modeling of cross-agent dependencies.  
Experiments on diverse knowledge-intensive QA benchmarks demonstrate the effectiveness and efficiency of our approach.

\section*{Limitations}
Despite the promising results achieved by \emph{AMATA}, our work has several limitations that warrant further investigation. Due to constraints in computational resources, our experiments were conducted primarily on a 7B-parameter model. We anticipate that scaling up to larger models (e.g., 70B parameters or beyond) could further enhance performance, particularly for more complex knowledge-intensive tasks.
Additionally, the number of winning and losing samples used in our dependency-aware DPO was limited to $M = N = 10$, which may not fully capture the diversity of inter-agent dependencies in more heterogeneous task settings. We set the Top-$K$ value to 5 for ranking winning examples, a conservative choice that balances efficiency and effectiveness but may overlook finer-grained preference structures. Future work will explore larger sample sizes and more adaptive ranking strategies as computational capacity increases.

\section*{Acknowledgments}
This work was supported by the National Natural Science Foundation of China (Grant No. 62506110). It was also supported by the Natural Science Foundation of Anhui Province, China (Grant No. 2508085QF227) and the Hefei University of Technology Scientific Research Innovation Start-up Special Project Type A (Grant No. JZ2025HGQA0137).

\bibliography{custom}
\bibliographystyle{acl_natbib}

\appendix

\section{Detailed Experimental Settings}
\label{experimental_res}

\subsection{Datasets and Evaluation Metrics}
\label{data_metrics}

\subsubsection{Trajectory Training Set Construction}
\begin{figure}[!t]
  \centering
  \includegraphics[height=8cm, width=7cm]{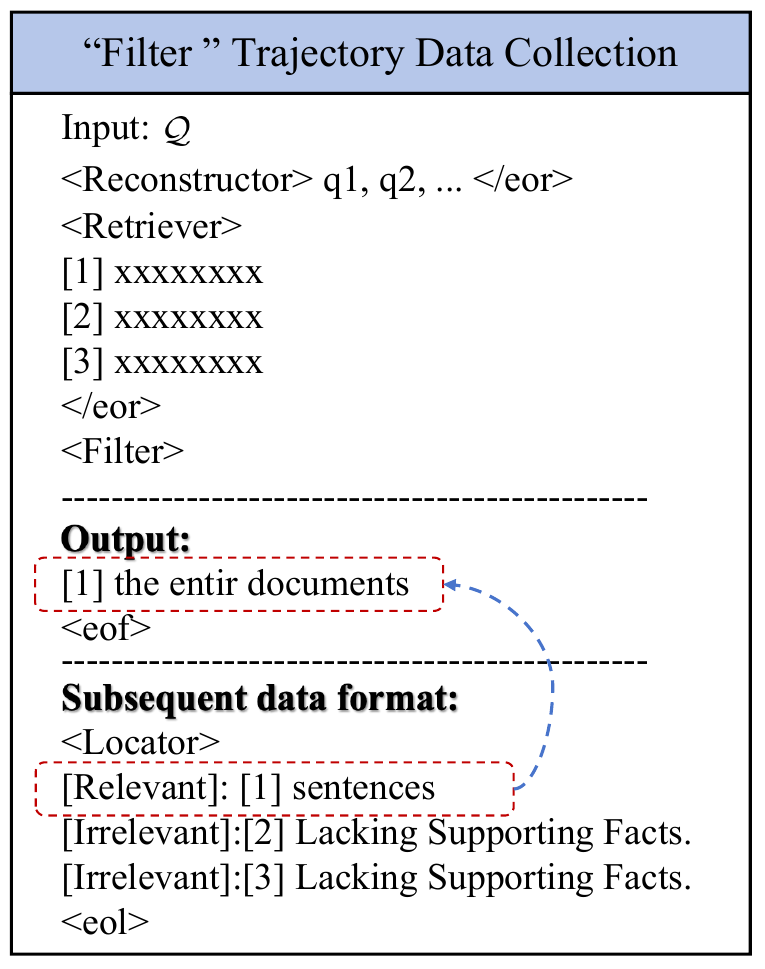}
  \caption{Collection of ``Filter'' trajectory data.}
  \label{filter_trajectory}
\end{figure}

\begin{figure}[!t]
  \centering
  \includegraphics[height=9cm, width=7cm]{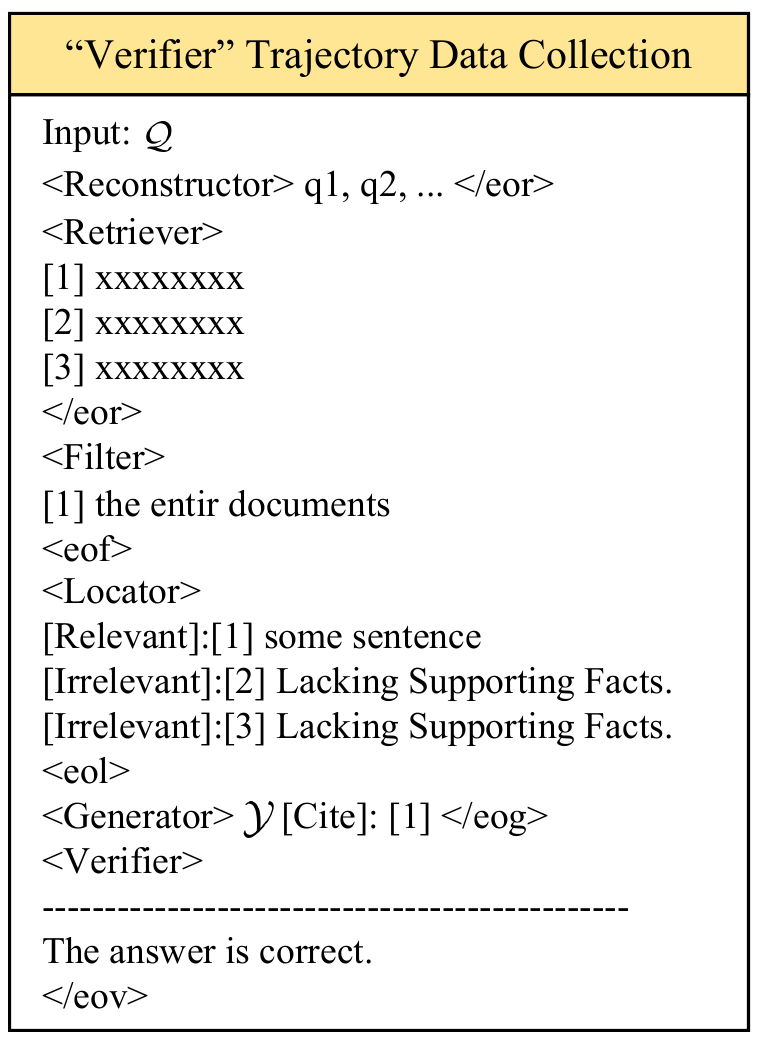}
  \caption{Collection of ``Verifier'' trajectory data.}
  \label{verifier_trajectory}
\end{figure}

Our trajectory data are collected from open-source long-trajectory datasets provided by Self-RAG~\cite{DBLP:conf/iclr/AsaiWWSH24} and SMART~\cite{DBLP:conf/aaai/YueWCHW25}, which together include 140{,}000 well-designed instances.\footnote{The open-source trajectory data are available at \url{https://huggingface.co/datasets/ShengbinYue/Long-short-Trajectory}.}
Trajectory data for two additional agents are derived from the previously constructed basic trajectories. Figure~\ref{filter_trajectory} illustrates the collection process for \texttt{$\langle$Filter$\rangle$} trajectory data, which is guided by the \texttt{$\langle$Retriever$\rangle$} step and subject to subsequent \texttt{$\langle$Locator$\rangle$} constraints. Data collection for the \texttt{$\langle$Verifier$\rangle$} trajectory is performed after the \texttt{$\langle$Generator$\rangle$} step, and aims to verify answer robustness as shown in Figure~\ref{verifier_trajectory}.

\begin{table*}[!t]
\centering
\begin{tabular}{ccc}
\toprule
\textbf{Agent Roles}  & High Score (4--5) Condition & Low Score (0--1) Condition \\ \midrule
\textbf{Retriever/Filter/Locator}    & \makecell[l]{Question requires external,\\ long-tail, or specific factual\\ knowledge.}   & \makecell[l]{Question can be answered with\\ the LLM's parametric\\ knowledge alone.} \\ \midrule
\textbf{Verifier} & \makecell[l]{The answer is complex, potentially\\ ambiguous, or requires high factual\\ precision.} & \makecell[l]{The answer is straightforward\\ or the confidence from previous\\ steps is very high.} \\ \midrule
\textbf{Reconstructor} & \makecell[l]{The user instruction is complex,\\ multi-hop, or requires semantic\\ parsing.} & \makecell[l]{The question is already\\ simple and well-structured.} \\ \midrule
\textbf{Generator}   & Always required (baseline score of 5). & N/A \\ \bottomrule
\end{tabular}
\caption{Detailed rubric for different agent roles in preference scoring.}
\label{detailed_rubric_score}
\end{table*}

\begin{table}[!tb]
\centering
\small
\begin{tabular}{ccc}
\toprule
\textbf{Group}  & \textbf{Knowledge Agent Scores} & \textbf{PopQA Accuracy} \\ \midrule
Group 1 & 1.2 & 22.5\%  \\ \midrule
Group 2 & 3.1 & 41.8\%  \\ \midrule
Group 3 & 4.2 & 63.4\%  \\ \bottomrule
\end{tabular}
\caption{Correlation analysis of the reasonableness of LLM-generated preferences. ``Group~1'', ``Group~2'', and ``Group~3'' correspond to low, medium, and high knowledge requirements, respectively.}
\label{correlation_analysis}
\end{table}

\subsubsection{Trajectory Scores}
\label{trajectory_scores_app}
\begin{figure}[!t]
  \centering
  \includegraphics[width=8cm]{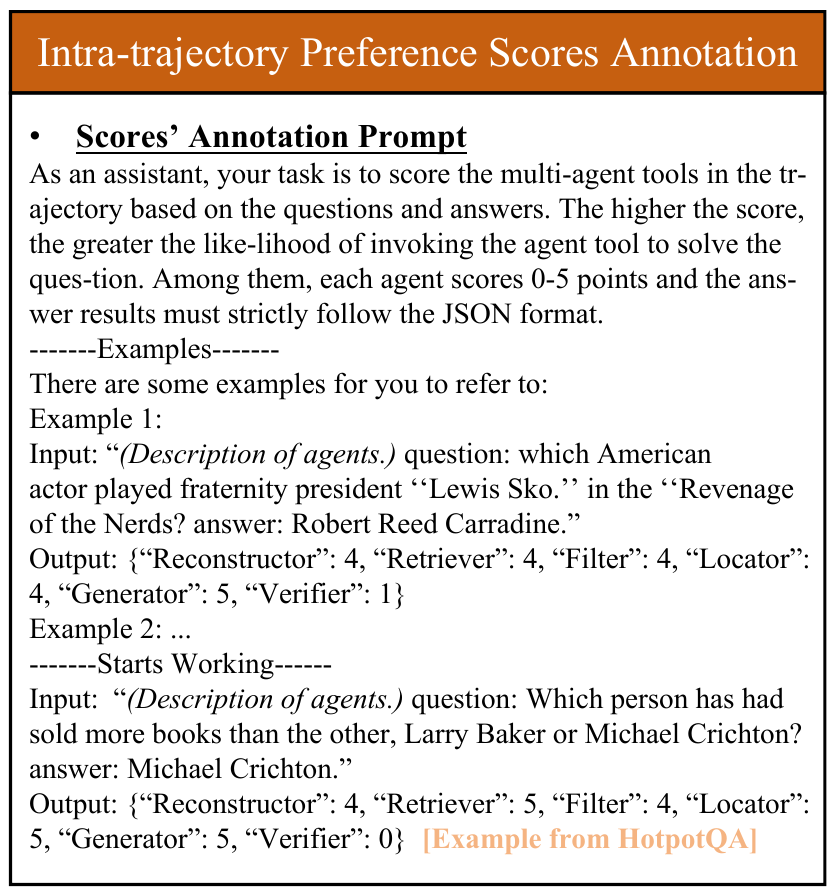}
  \caption{Intra-trajectory score collection.}
  \label{intra_scores}
\end{figure}

\begin{figure*}[!t]
  \centering
  \includegraphics[width=16cm]{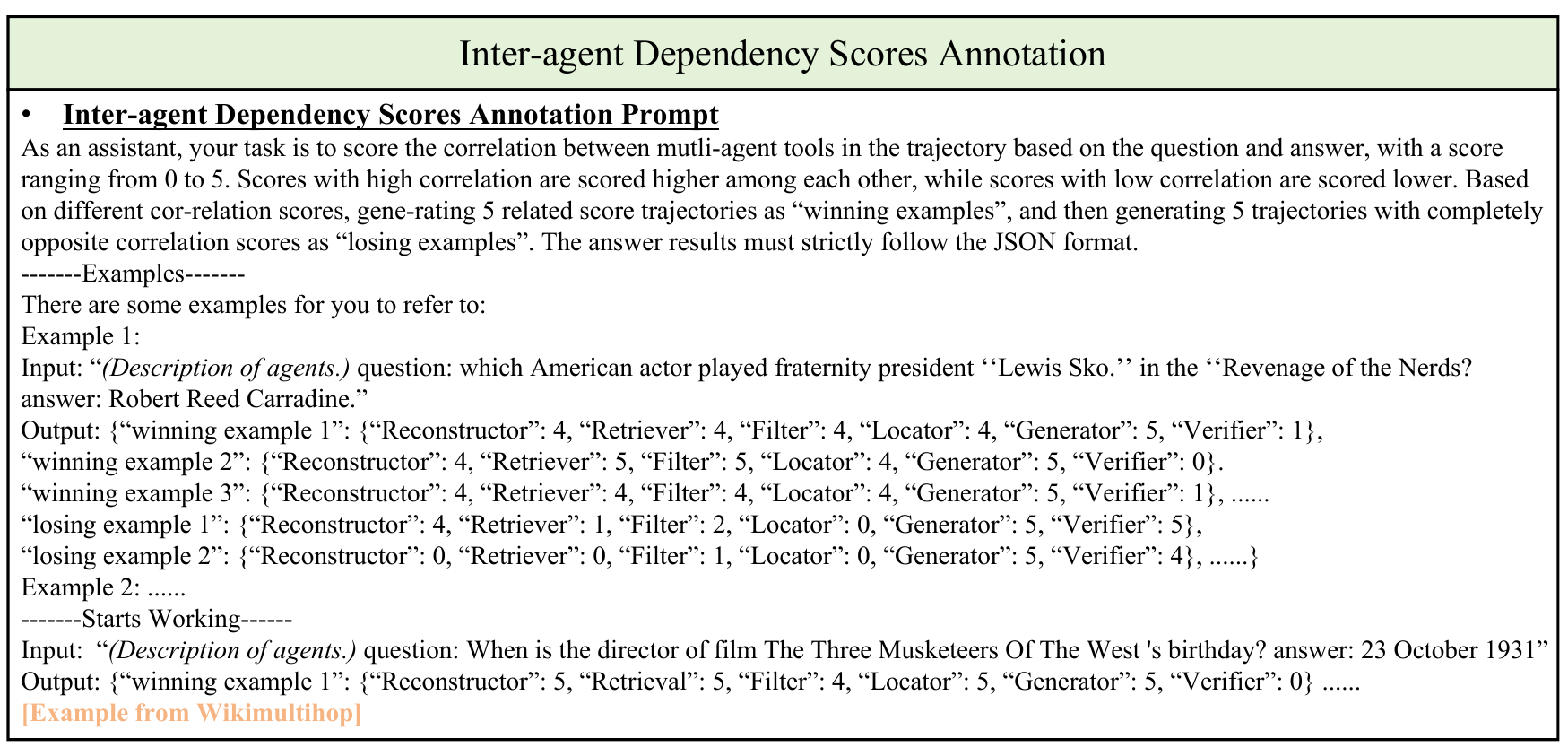}
  \caption{Inter-trajectory score collection.}
  \label{inter_scores}
\end{figure*}

\begin{figure*}[!t]
  \centering
  \includegraphics[width=16cm]{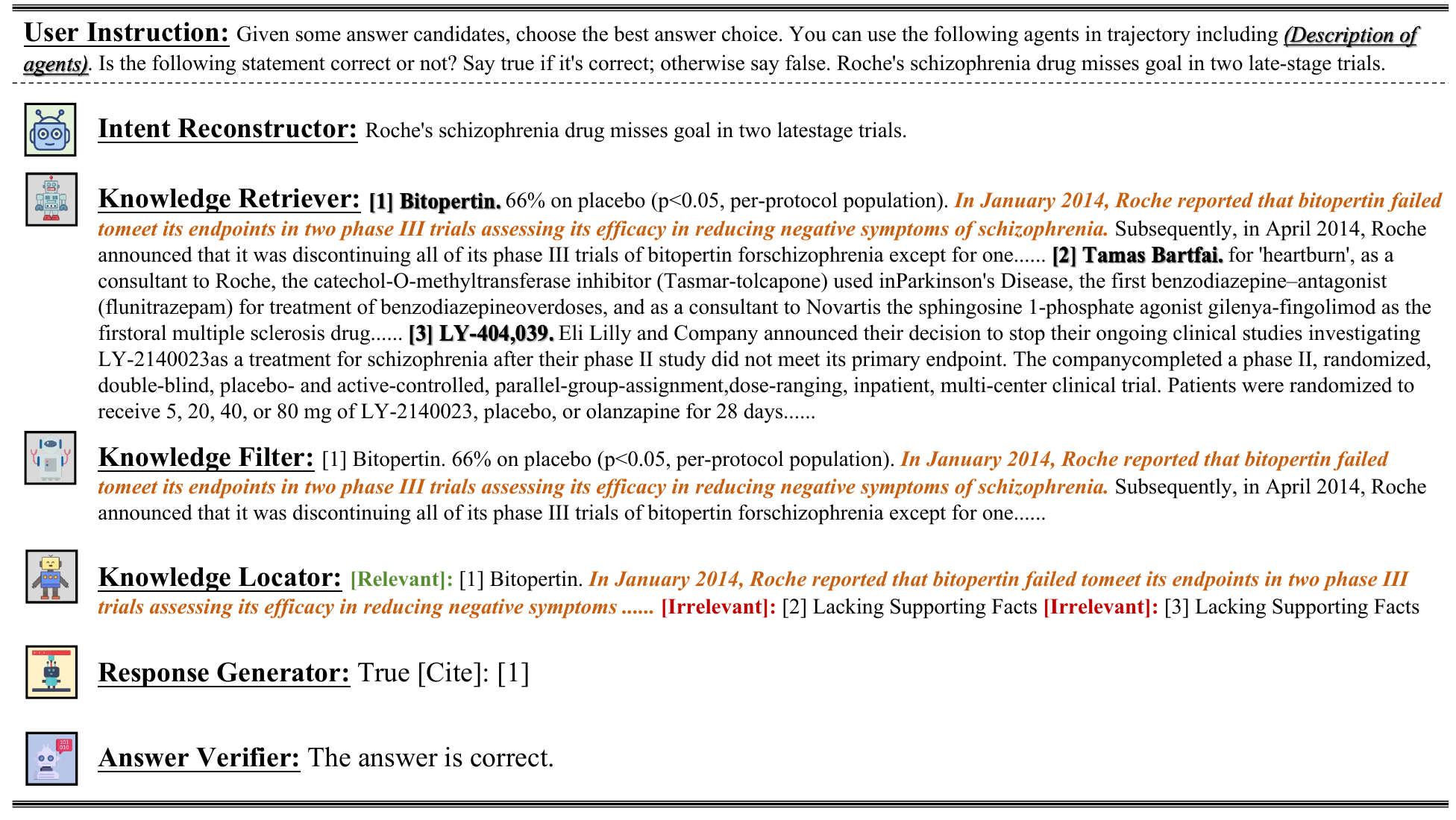}
  \caption{Complete response example from PubHealth.}
  \label{complete_example}
\end{figure*}

As shown in Figure~\ref{intra_scores}, intra-trajectory scores are computed based on QA pairs and two demonstration examples. Additionally, inter-trajectory score data are annotated using demonstration examples with varying task instructions, as depicted in Figure~\ref{inter_scores}. A comprehensive training example is provided in Figure~\ref{complete_example}.
Inter-dependency preference data are sorted based on the sum of scores across trajectories.

There are two primary types of agents involved in scoring: knowledge agents, and generator/verifier agents.
Knowledge agents are essential when multi-agent systems require external knowledge to solve complex questions.
Generator/verifier agents are associated with the confidence level of LLMs during answer generation.

To better elucidate the rationality of our scoring process, we provide a detailed rubric for agent preference scoring in Table~\ref{detailed_rubric_score}.
For example, for the question ``Which American actor played fraternity president ``Lewis Skol.'' in the ``Revenge of the Nerds'' comedy films?'' (see Figure~\ref{instruction_example}), the LLM annotator correctly assigns high scores to the knowledge agents (\texttt{Retriever=5}, \texttt{Filter=4}, \texttt{Locator=4}) as it identifies the need for specific external knowledge, and assigns a low score to the Verifier due to high confidence in the retrieved evidence.

\begin{figure*}[!t]
\centering
\includegraphics[width=16cm]{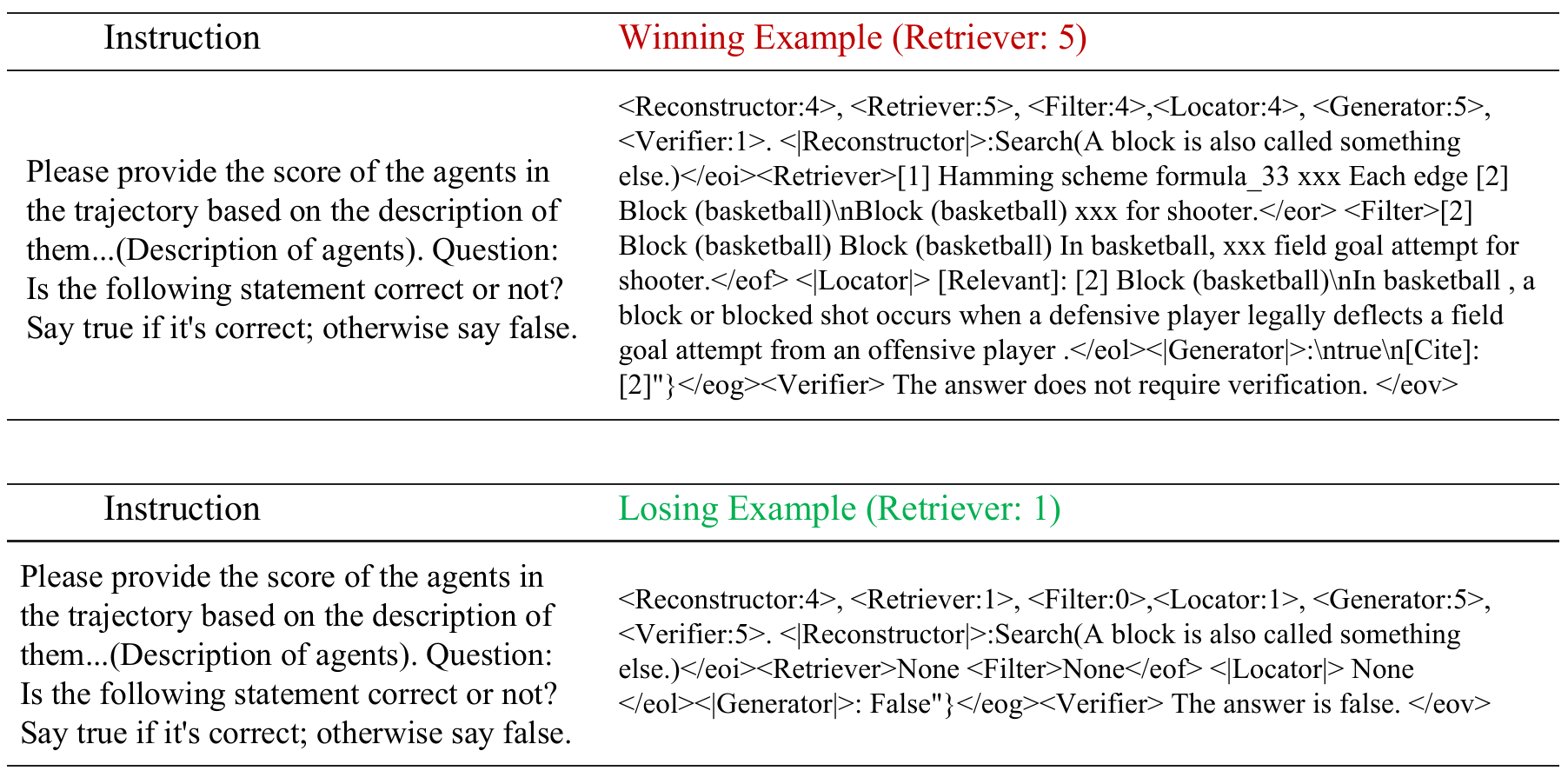}
\caption{Examples of different preference scores for \texttt{$\langle$Retriever$\rangle$}.}
\label{dpo_scores}
\end{figure*}

We verify the reasonableness of these scores through two approaches: ablation studies and correlation analysis.
\begin{itemize}
    \item \textbf{Ablation Study as Direct Evidence:} The most direct validation is to observe performance changes when removing an agent deemed important according to the score. Our ablation study (Table~\ref{ablation_study}) provides strong evidence that the LLM-assigned preferences align with the agents' actual functional importance. For example, on PopQA (a long-tail knowledge dataset), removing $\mathcal{A}_{\text{KR}}$ (typically high-scored) results in a $\sim$3.1\% performance drop, the largest among single-agent ablations. This confirms that the LLM annotator correctly identifies the critical need for retrieval in such tasks. Conversely, removing $\mathcal{A}_{\text{AV}}$ has a smaller but still notable impact, particularly on factuality-focused tasks like ARC-C and ASQA, justifying its medium-to-low but non-zero scores.

    \item \textbf{Correlation Analysis:} To further support our claim, we conduct a correlation analysis on a sample of 200 questions from the PopQA dataset, comparing the average preference score assigned to the three knowledge agents (Retriever, Filter, Locator) against final task accuracy. As shown in Table~\ref{correlation_analysis}, the strong positive correlation demonstrates that higher LLM-assigned preference scores for knowledge agents correspond to significantly improved accuracy, thus quantitatively confirming the reasonableness of the scores.

    \item \textbf{DPO Score Examples:} To aid the LLM in understanding our scoring process, we provide two examples as demonstrations in the API input prompts, shown in Figure~\ref{dpo_scores}.
\end{itemize}

In summary, while annotation is performed by an LLM, it is grounded in a structured, semantically meaningful rubric. More importantly, its effectiveness is empirically confirmed through ablation studies and correlation analysis, demonstrating that the learned preferences contribute directly to the framework's performance and efficiency.

\subsubsection{Evaluation Datasets and Metrics}
\begin{itemize}
    \item \textbf{Fact Verification:} PubHealth (also referred to as HealthQA)~\cite{pubhealthtab} is a public health fact-checking dataset. Model performance is evaluated by accuracy (Acc.) on its test set of 987 samples labeled ``True'' or ``False''.
    \item \textbf{Multiple-choice QA:} ARC-Challenge~\cite{DBLP:journals/corr/abs-1803-05457} consists of 1,172 multiple-choice science exam questions. Performance is measured by accuracy (Acc.).  
    \item \textbf{Open-domain QA:} (1) PopQA~\cite{DBLP:journals/corr/abs-2212-10511} contains 1,399 long-tail, rare-entity queries from Wikipedia. (2) SQuAD~\cite{DBLP:conf/emnlp/RajpurkarZLL16} includes 8,886 queries written by annotators based on documents. Following prior work~\cite{DBLP:conf/iclr/AsaiWWSH24}, performance is evaluated using exact match (EM).  
    \item \textbf{Ambiguous QA:} ASQA~\cite{DBLP:conf/emnlp/GaoYYC23} features 4,132 ambiguous factual questions requiring long-form responses. Fluency is assessed using Mauve, and accuracy is measured with Str\_EM and Rouge-L, consistent with official evaluation settings.
\end{itemize}

\subsection{Baselines}

\subsubsection{Vanilla QA Methods}
LLMs acquire extensive factual knowledge, internalized within their model parameters through large-scale unsupervised pre-training. During both training and inference, we adhere to official prompt formats.
\begin{itemize}
    \item \textbf{SFT and preference alignment models:} GPT-3.5-turbo (ChatGPT)\footnote{We use the gpt-3.5-turbo-0125 version in experiments.}, Llama-2-Chat-7B~\cite{DBLP:journals/corr/abs-2302-13971}, and Llama-2-Chat-13B~\cite{DBLP:journals/corr/abs-2302-13971}.
    \item \textbf{SFT models:} Instruct-v0.2-7B~\cite{DBLP:journals/corr/abs-2310-06825}, Vicuna-v1.5-13B~\cite{DBLP:conf/nips/ZhengC00WZL0LXZ23}, Qwen2.5-7B-Instruct~\cite{qwen25}, and Alpaca2-7B\footnote{\url{https://github.com/tatsu-lab/stanford_alpaca}}.
\end{itemize}

\subsubsection{Knowledge-Augmented Methods}
We implement standard knowledge augmentation approaches. When model weights are unavailable, methods are replicated using the same base models and training data. Uniform retrieval models and knowledge bases ensure experimental fairness.
\begin{itemize}
    \item \textbf{REPLUG}~\cite{DBLP:conf/naacl/ShiMYS0LZY24} uses frozen LLM parameters and augments them with a tunable retrieval model. In our experiments, the backbone is replaced with Llama-2-Chat-7B for fairness.
    \item \textbf{VANILLA-7B}~\cite{DBLP:conf/emnlp/GaoYYC23} first retrieves relevant passages, then instructs the model to assess document relevance and generate appropriate citations. The backbone is Llama-2-Chat-7B.
    \item \textbf{INTERACT-7B}~\cite{DBLP:conf/emnlp/GaoYYC23} employs an interactive prompting mechanism that enables the agent to verify retrieved passages through three distinct actions: ``Check'', ``Output'', and ``End''. The backbone is Llama-2-Chat-7B.
    \item \textbf{RADIT-7B}~\cite{DBLP:conf/iclr/Lin0CSL00KSLZY24} introduces retrieval-augmented dual instruction tuning, a lightweight fine-tuning framework that retrofits existing LLMs with retrieval capabilities, offering an alternative to conventional methodologies. For fair comparison, the pre-trained Llama-2 model is fine-tuned on the same dataset used in our experiments.
    \item \textbf{SelfRAG-7B}~\cite{DBLP:conf/iclr/AsaiWWSH24} enhances language model quality and factuality through retrieval and self-reflection with special tokens.
\end{itemize}

\subsubsection{LLM-Based Trajectory Methods}
Through multi-agent collaboration, LLMs with distinct task capabilities can be coordinated to form workflows that enhance response reliability.
\begin{itemize}
    \item \textbf{MMAgent-3*7B:} Our modular multi-agent framework, in which each component agent is independently trained on identical datasets while sharing the same pre-trained Llama-2 backbone. The workflow is realized through systematic agent decoupling.
    \item \textbf{SMART}~\cite{DBLP:conf/aaai/YueWCHW25}: A global-local multi-agent framework with predefined trajectories.
    \item \textbf{GiGPO}~\cite{DBLP:journals/corr/abs-2505-10978}: Proposes a hierarchical architecture that evaluates both global trajectory quality and local action effectiveness, while eliminating the need for auxiliary models or additional rollouts. This paradigm ensures superior scalability for long-horizon LLM agent training.
    \item \textbf{SPA-RL}~\cite{DBLP:journals/corr/abs-2505-20732}: Proposes a general reward redistribution framework that systematically decomposes the final reward into stepwise contributions, with each component accurately reflecting its incremental impact on overall task completion.
\end{itemize}

\subsection{Experimental Settings}

\subsubsection{Training Details}
\label{training_details}
Our implementation is initialized with the pre-trained Llama-2-7B foundation model~\cite{DBLP:journals/corr/abs-2302-13971}.
Each agent is initialized as an independent large model, emulating distinct agent abilities through special tokens in both intra- and inter-trajectory training. These abilities are acquired by training a shared model~\cite{DBLP:conf/iclr/AsaiWWSH24,DBLP:conf/aaai/YueWCHW25,DBLP:conf/acl/QiaoQRWR00JX0C25}.
Training is performed on two NVIDIA A100 GPUs (80GB each), utilizing LoRA adaptation~\cite{DBLP:conf/iclr/HuSWALWWC22} for intra- and inter-trajectory learning.
Both trajectory learning phases span three epochs with uniform hyperparameters: batch size of 64, peak learning rate of $6 \times 10^{-4}$, and 5\% warmup steps.
The maximum sequence length is 1024 tokens for intra-trajectory and 2048 tokens for inter-trajectory training, with DeepSpeed Stage-3~\cite{Ruwase_He_2020} applied to optimize GPU memory utilization.

Due to the advantage of supervision at intermediate agent training steps, RL-based trajectory methods such as GiGPO~\cite{DBLP:journals/corr/abs-2505-10978} and SPA-RL~\cite{DBLP:journals/corr/abs-2505-20732} outperform simple multi-agent baselines.
Specifically, we assign intermediate steps in the GiGPO trajectory to the six multi-agent actions we define.
The final reward signal for correctly predicted answers is set to 1, and to $-1$ for incorrect answers.
GiGPO obtains intermediate reward signals by implicitly learning episode-relative advantages and thus does not require manual specification in the task dataset.
SPA-RL, however, necessitates explicit process supervision signals at intermediate steps~\cite{DBLP:journals/corr/abs-2505-20732}.
Here, we utilize GPT-4o~\cite{gpt4_} to score each step, providing intermediate human feedback as accurate as possible. The final reward setting matches that used for GiGPO.

\subsubsection{Evaluation Details}
For the two additional agents—knowledge filter (\texttt{$\langle$Filter$\rangle$}) and verifier (\texttt{$\langle$Verifier$\rangle$})—if \texttt{$\langle$Filter$\rangle$} outputs retrieved document indices not present in \texttt{$\langle$Retriever$\rangle$}, we remove them. If all indices are filtered out, we retain the first one.
If the output trajectory includes \texttt{$\langle$Verifier$\rangle$}, the answer requires further verification.
If the \texttt{$\langle$Verifier$\rangle$} output is ``wrong'', we incorporate the error signal into the instruction via the prompt and re-execute the trajectory for LLM reflection.
Otherwise, we extract the answer from \texttt{$\langle$Generator$\rangle$}.
Other settings, including evaluation task instructions, follow those of the SMART model~\cite{DBLP:conf/aaai/YueWCHW25}.

\subsection{Agent Description}
\label{agent_desc_sec}
We extend the SMART~\cite{DBLP:conf/aaai/YueWCHW25} framework by adding two essential agents to further enhance the reliability of LLM responses: the Knowledge Filter and Answer Verifier.

\begin{itemize}
    \item \textbf{Intent Reconstructor:} Agent $\mathcal{A}_{\text{IR}}$ elucidates user question intent. To process diverse instructions into well-formatted intents, the agent employs four key capabilities: (1) integrating contextual clues, (2) identifying key queries, (3) unifying task formulation, and (4) decomposing intent.
    \item \textbf{Knowledge Retriever:} Given the reconstructed question, agent $\mathcal{A}_{\text{KR}}$ retrieves supplementary knowledge from an external knowledge base (e.g., Wikipedia). For simple questions, \emph{AMATA} may skip intermediate knowledge agents (i.e., $\mathcal{A}_{\text{KR}}$, $\mathcal{A}_{\text{KF}}$, and $\mathcal{A}_{\text{KL}}$) and directly invoke the response generator $\mathcal{A}_{\text{RG}}$.
    \item \textbf{Knowledge Filter:} To eliminate redundant information from retrieved documents, agent $\mathcal{A}_{\text{KF}}$ extracts the most accurate background knowledge. Concurrently, $\mathcal{A}_{\text{KF}}$ empowers the locator agent $\mathcal{A}_{\text{KL}}$ to constrain the search scope and perform fine-grained identification of knowledge supportive to the generator $\mathcal{A}_{\text{RG}}$.
    \item \textbf{Knowledge Locator:} Operating on the refined document set from $\mathcal{A}_{\text{KF}}$, agent $\mathcal{A}_{\text{KL}}$ performs granular localization to identify and extract knowledge segments most conducive to response generation by $\mathcal{A}_{\text{RG}}$.
    \item \textbf{Response Generator:} Agent $\mathcal{A}_{\text{RG}}$ synthesizes responses in two modes: when knowledge segments are supplied by $\mathcal{A}_{\text{KL}}$, it generates answers constrained by localized evidence with explicit source attribution; otherwise, it relies solely on parametric knowledge. The generator maintains provenance transparency by clearly differentiating evidence-based and intrinsic knowledge sources.
    \item \textbf{Answer Verifier:} Agent $\mathcal{A}_{\text{AV}}$ performs self-correction by re-examining the generated response against relevant knowledge sources, identifying inaccuracies through evidence-based verification, and applying targeted revisions to enhance factual consistency and logical coherence.
\end{itemize}

We further detail the steps for the relevant agents (i.e., $\mathcal{A}_{\text{IR}}$, $\mathcal{A}_{\text{KF}}$, and $\mathcal{A}_{\text{AV}}$) to clarify how specific agent trajectory data are constructed.
For other agents' data collection steps, refer to Self-RAG~\cite{DBLP:conf/iclr/AsaiWWSH24}.

\begin{itemize}
    \item \textbf{Intent Reconstructor:} Within multi-turn dialogues, this agent models dependencies to capture long-term intent. When processing noisy instructions, it eliminates extraneous content to isolate essential questions. For diverse task formats (e.g., multiple-choice QA), the agent standardizes inputs into a cohesive query representation. For complex multi-hop questions (e.g., \texttt{``Who was born earlier, person A or person B?''}), it decomposes them into atomic intents, such as retrieving each individual's birthdate. By flexibly applying these capabilities, the agent derives a well-structured query intent, facilitating external knowledge retrieval.
    \item \textbf{Knowledge Filter:} Prompt-induced hallucinations may cause the filter $\mathcal{A}_{\text{KF}}$ to generate documents and indices not present in the retriever $\mathcal{A}_{\text{KR}}$. To address this, we implement rigorous content verification procedures to ensure filtered content remains consistent with original sources. Additionally, if filtering removes all documents, the top-ranked document based on the retriever's scores is retained.
    \item \textbf{Answer Verifier:} For responses marked ``correct'' by agent $\mathcal{A}_{\text{AV}}$, the reasoning process for that question is terminated. Conversely, for responses marked ``wrong,'' error signals are incorporated into the instruction $\mathcal{Q}$ via concatenation. This encourages the multi-agent trajectory to adopt a reasoning mode that considers potential errors, prompting deeper reflection and improving the model's reasoning accuracy.
\end{itemize}

\section{Additional Experimental Results}
\label{additional_results}

\begin{figure*}[!tb]
  \centering
  \includegraphics[height=6cm]{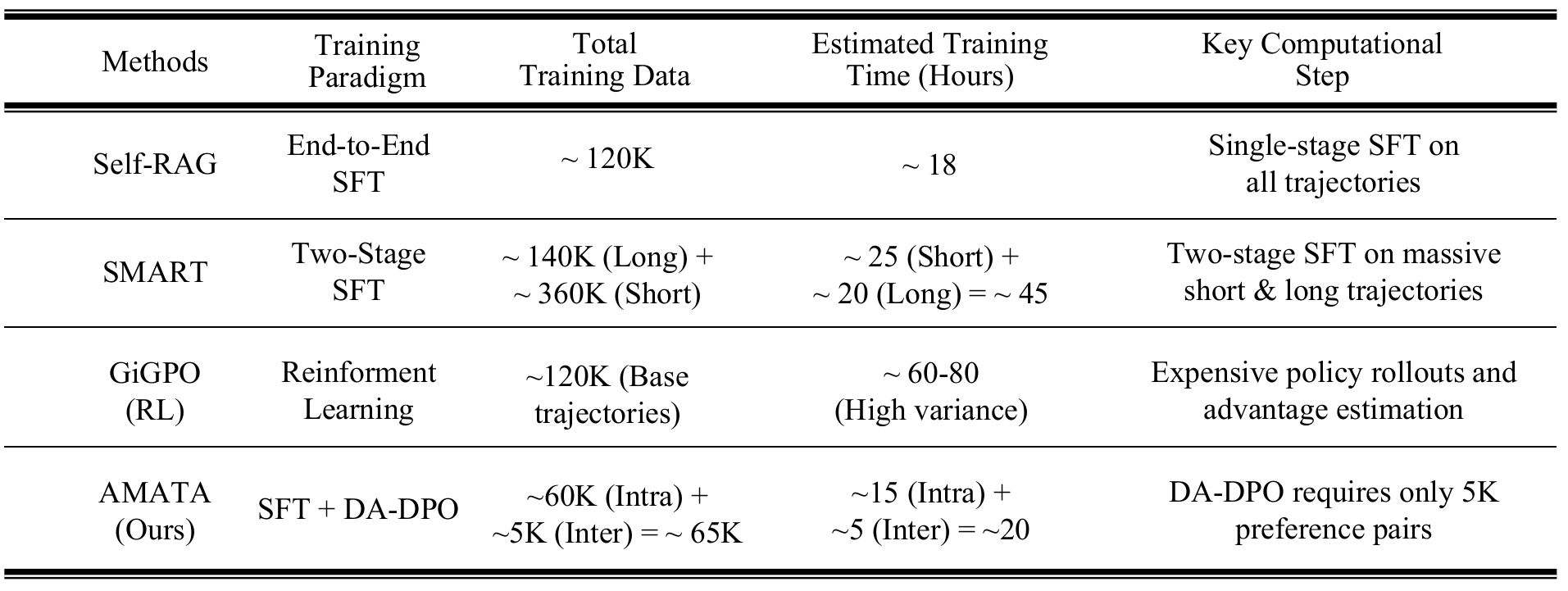}
  \caption{Comparison of training data size and computational cost for various baselines and our \emph{AMATA} model.}
  \label{datasize_comp}
\end{figure*}

\subsection{Data Size and Computational Cost Comparison}
\label{datasize_comp_app}

In Figure~\ref{datasize_comp}, we present a detailed comparison of the training data size and computational cost for \emph{AMATA} and its key baselines. All experiments are performed using identical hardware settings (2 $\times$ NVIDIA A100 80GB GPUs) to ensure fairness.

We observe the following:  
(1) \textbf{Data Efficiency:} The total data consumption of \emph{AMATA} ($\sim$65{,}000 samples) is substantially lower than that of SMART ($\sim$500{,}000 samples) and is comparable to SelfRAG. Most importantly, the proposed DA-DPO stage is highly data-efficient, requiring only $\sim$5,000 expert-ranked samples to effectively learn complex inter-agent dependencies. This is a fraction of the data used by other training stages and baselines.  

(2) \textbf{Computational and Training Time Efficiency:}  
[1] Compared to SMART, \emph{AMATA} requires less than half the training time ($\sim$20 hours vs. $\sim$45 hours), mainly because pre-training on a massive, separate short-trajectory dataset is unnecessary. Our intra-trajectory learning consolidates this into a single, more efficient stage.  
[2] Compared to RL methods (GiGPO), \emph{AMATA} is 3--4 times faster. RL training is notoriously slow due to multiple rollouts and per-step reward computation, whereas our DA-DPO stage performs stable, offline optimization.  
[3] The additional cost of DA-DPO over standard DPO is minimal (an extra $\sim$5 hours), as it utilizes the same computational framework but employs a more sophisticated loss function.

\begin{figure}[!tb]
  \centering
  \includegraphics[height=6cm]{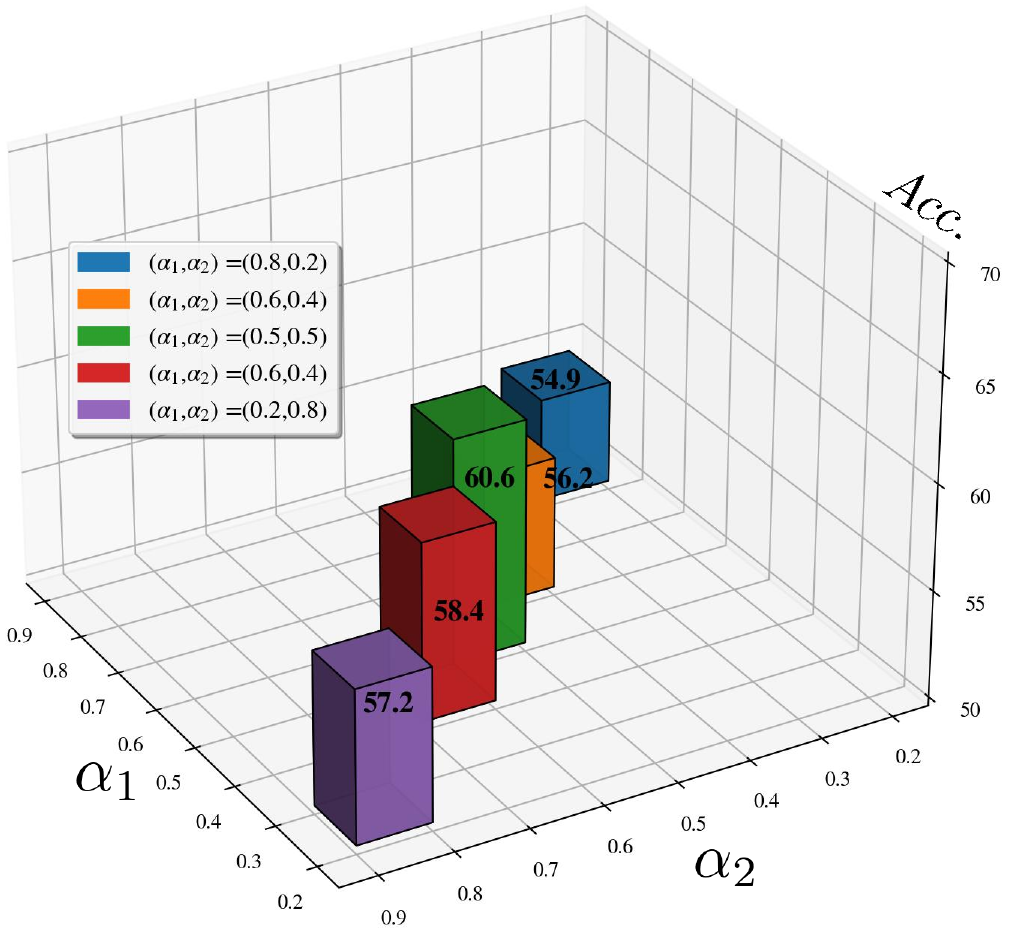}
  \caption{Averaged performance of \emph{AMATA} corresponding to different loss coefficients $\alpha_1$ and $\alpha_2$.}
  \label{alpha12}
\end{figure}

\begin{figure}[!t]
  \centering
  \includegraphics[height=6cm]{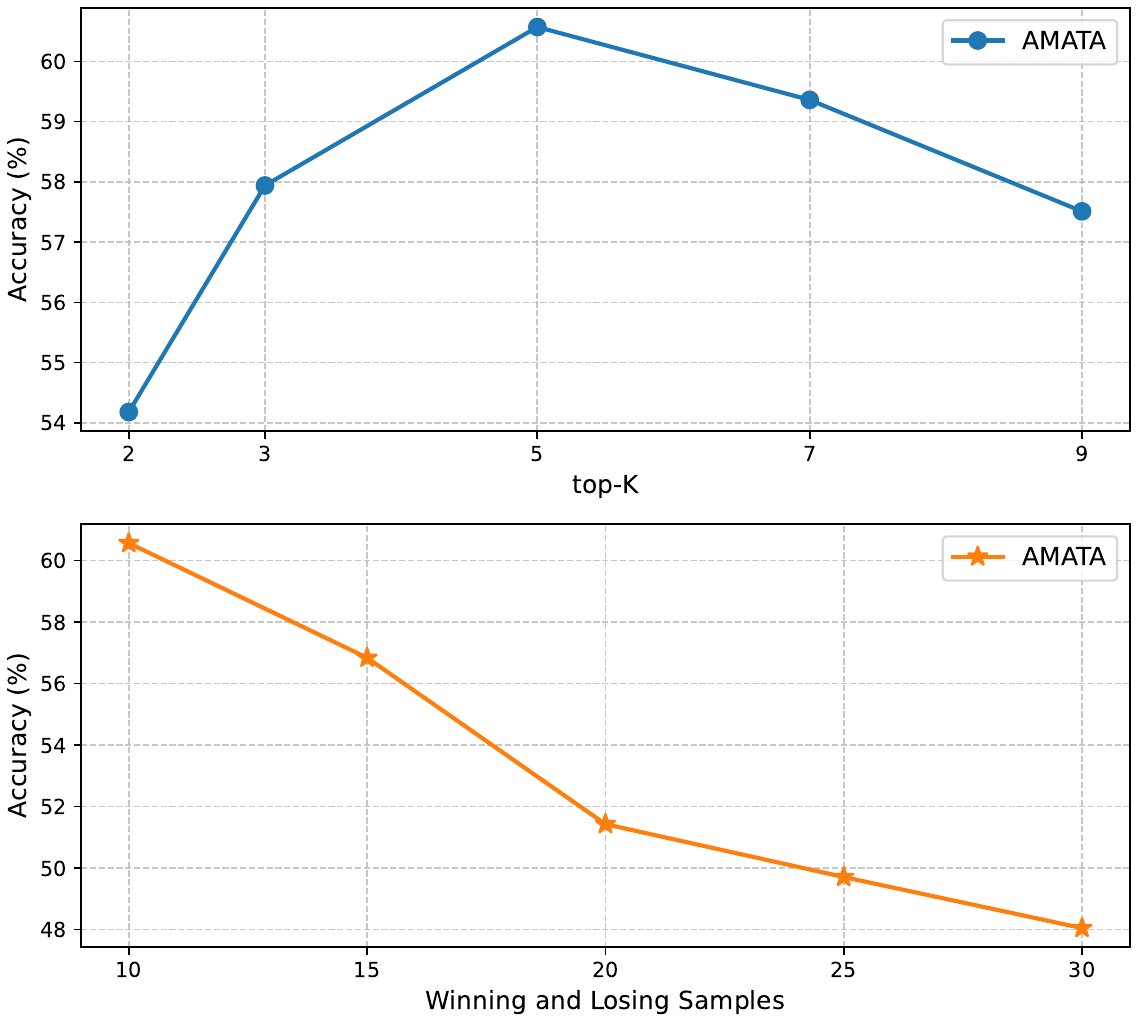}
  \caption{Impact of selecting the $K$ value for winning and losing samples on model performance.}
  \label{top_k_winninglosing}

\end{figure}

\begin{figure}[!tb]
  \centering
  \includegraphics[height=7cm, width=8cm]{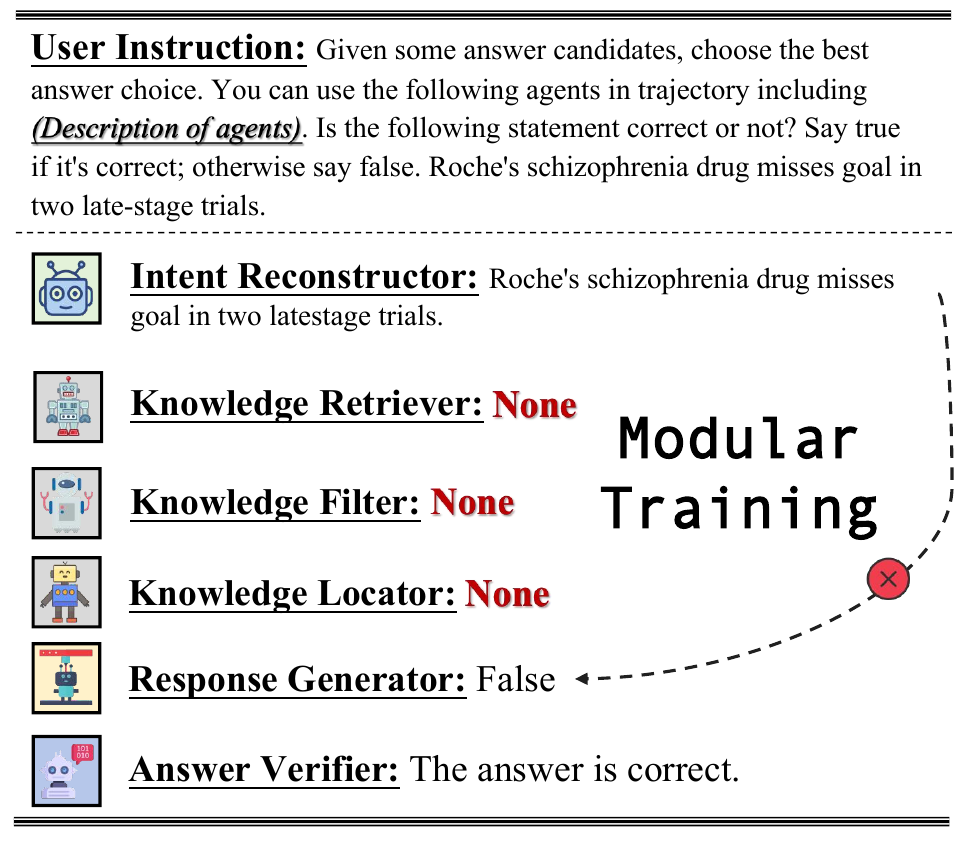}
  \caption{Complete response example from MMAgent.}
  \label{case_study1}
\end{figure}

\begin{table*}[!tb]
\centering
\footnotesize
\setlength{\tabcolsep}{3pt}
    \begin{tabular}{lccccccccc}
      \toprule
      \multicolumn{1}{c}{\textbf{Task} $\rightarrow$}  & \textbf{HealthQA} & \textbf{ARC-C} & \textbf{PopQA} & \textbf{Squad1} & \multicolumn{3}{c}{\textbf{ASQA}} & \multirow{2}{*}{\textbf{Average}} \\  
       \multicolumn{1}{c}{\textbf{Model} $\downarrow$}  & Acc. & Acc. & Acc. & Acc. & Str\_EM & Rouge-L & Mauve & \\  \midrule
\multicolumn{1}{c}{ $\text{Llama-3-Ins.}\text{8B}$} & 64.39$_{(\pm0.9)}$ & 53.44$_{(\pm1.5)}$ & 22.73$_{(\pm1.2)}$ & 16.24$_{(\pm0.8)}$ & 18.85$_{(\pm1.7)}$ & 33.48$_{(\pm1.5)}$ & 37.65$_{(\pm0.7)}$ & 35.25$_{(\pm1.3)}$ \\
\multicolumn{1}{c}{\it $\text{GPT4o}$} & \it 79.24$_{(\pm1.0)}$ & \it 80.71$_{(\pm1.8)}$ & \it 30.94$_{(\pm1.1)}$ & \it 24.83$_{(\pm1.5)}$ & \it 44.02$_{(\pm0.9)}$ & \it 36.92$_{(\pm1.7)}$ & \it 47.53$_{(\pm1.6)}$ & \it 49.17$_{(\pm1.1)}$ \\
\multicolumn{1}{c}{$\text{RADIT}\text{ 8B}$} & 55.29$_{(\pm1.1)}$ & 64.88$_{(\pm1.0)}$ & 41.15$_{(\pm1.6)}$ & 24.97$_{(\pm1.1)}$ & 29.01$_{(\pm1.8)}$ & 17.22$_{(\pm1.2)}$ & 13.86$_{(\pm0.8)}$ & 35.20$_{(\pm1.7)}$ \\
\multicolumn{1}{c}{$\text{SelfRag}\text{ 8B}$} & 70.89$_{(\pm1.6)}$ & 68.14$_{(\pm1.2)}$ & 43.55$_{(\pm1.5)}$ & 26.06$_{(\pm0.8)}$ & 30.95$_{(\pm1.1)}$ & 35.98$_{(\pm1.2)}$ & 87.12$_{(\pm2.1)}$ & 51.81$_{(\pm1.2)}$ \\ 
\multicolumn{1}{c}{ $\text{SMART} \text{ 8B}$ } & 75.99$_{(\pm1.6)}$ & 72.81$_{(\pm0.6)}$ & 47.66$_{(\pm1.2)}$ & 33.05$_{(\pm1.4)}$ & 45.74$_{(\pm1.6)}$ & 44.95$_{(\pm1.2)}$ & 94.80$_{(\pm1.2)}$ & 59.29$_{(\pm0.8)}$ \\
\multicolumn{1}{c}{ $\text{SPA-RL} \text{ 8B}$} & {76.88}$_{(\pm1.7)}$ & {72.98}$_{(\pm1.3)}$ & {46.85}$_{(\pm1.3)}$ & \underline{34.51}$_{(\pm1.8)}$ & \underline{46.63}$_{(\pm1.2)}$ & {45.57}$_{(\pm0.7)}$ & {94.93}$_{(\pm1.1)}$& 59.76$_{(\pm1.1)}$ \\
\multicolumn{1}{c}{ $\text{GiGPO} \text{ 8B}$} & \underline{77.20}$_{(\pm1.0)}$ & \underline{73.84}$_{(\pm0.7)}$ & \underline{47.29}$_{(\pm1.7)}$ & {34.05}$_{(\pm1.2)}$ & {45.75}$_{(\pm1.8)}$ & \underline{46.98}$_{(\pm1.0)}$ & \underline{95.62}$_{(\pm1.7)}$ & \underline{60.10}$_{(\pm1.6)}$ \\
\multicolumn{1}{c}{$\text{AMATA} \text{ 8B}$} & \textbf{78.74}$_{(\pm1.2)}$ & \textbf{74.11}$_{(\pm1.0)}$ & \textbf{49.62}$_{(\pm1.6)}$ & \textbf{37.80}$_{(\pm1.2)}$ & \textbf{50.83}$_{(\pm1.5)}$ & \textbf{51.57}$_{(\pm1.1)}$ & \textbf{96.92}$_{(\pm1.8)}$ & \textbf{62.80$_{(\pm1.3)}$} \\
\bottomrule
\end{tabular}
\caption{Overall results of \emph{AMATA} with other backbone LLMs.}
\label{backbone_experiments}
\end{table*}

\subsection{Hyperparameter Sensitivity Analysis}

\subsubsection{Loss Coefficients}
We evaluate the impact of the loss coefficients in the total loss $\mathcal{L}_{\text{total}}$ by experimenting with five different pairs of values for $\alpha_1$ and $\alpha_2$.  
As shown in Figure~\ref{alpha12}, our model achieves optimal performance when the coefficients are balanced (i.e., $\alpha_1 = 0.5$ and $\alpha_2 = 0.5$).

\subsubsection{Number of Winning and Losing Samples}
In Figure~\ref{top_k_winninglosing}, we analyze the effect of varying the number of winning and losing samples, denoted $M$ and $N$, as well as the top-$K$ value in our dependency-aware DPO framework.  
We observe that increasing $M$ and $N$ causes model performance to gradually decline, likely due to the introduction of excessive retrieval noise.
Consequently, we set both $M$ and $N$ to 10.
With $M = N = 10$, we further investigate how different values of $K$ affect performance across winning examples.
Our findings indicate that choosing $K$ as the midpoint of $M$ yields optimal results.
Therefore, we set $K = 5$ to achieve the best performance reported in Table~\ref{main_res}.

\begin{figure*}[!tb]
  \centering
  \includegraphics[height=8cm]{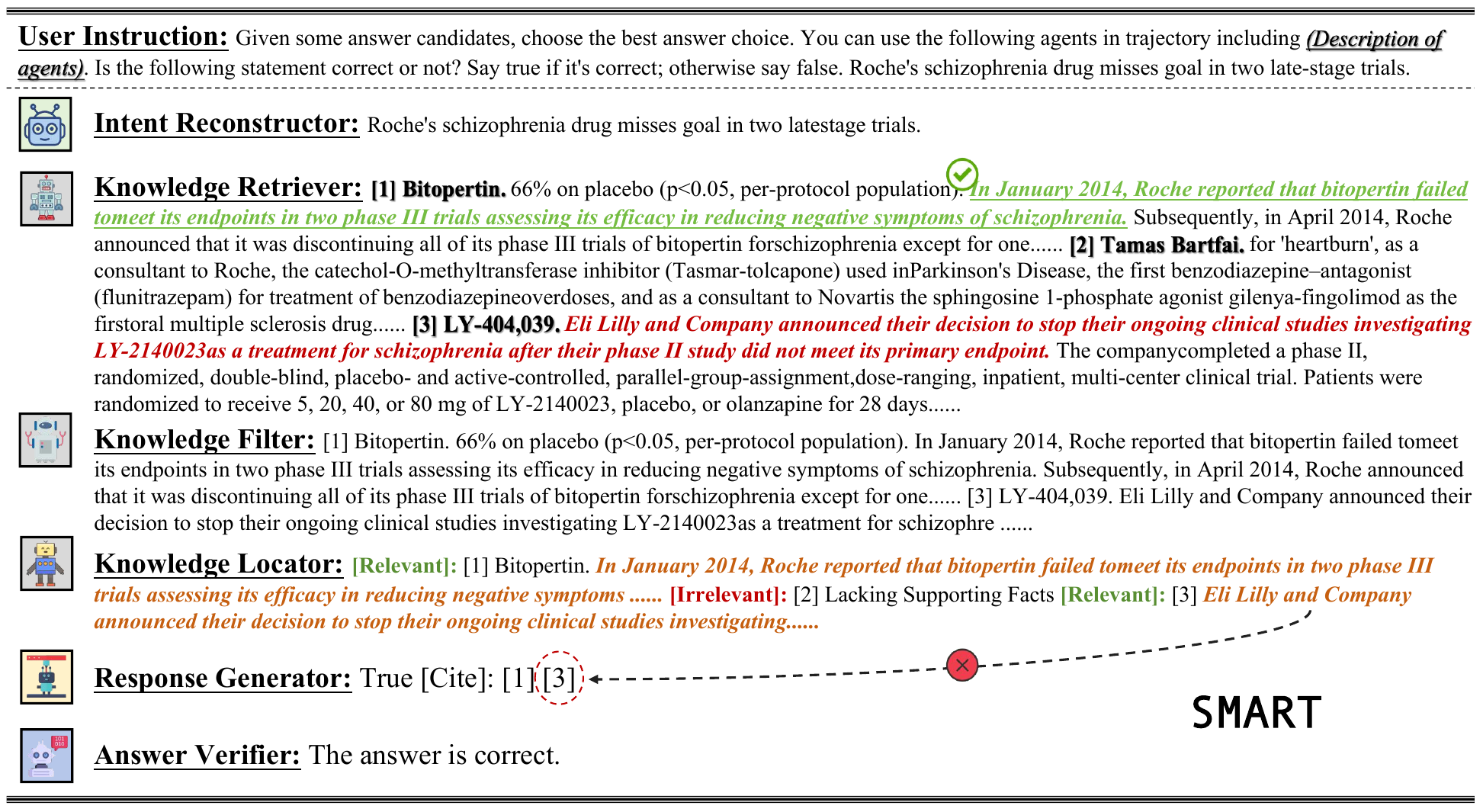}
  \caption{Complete response example from SMART.}
  \label{case_study2}
\end{figure*}

\begin{figure*}[!tb]
  \centering
  \includegraphics[height=8cm]{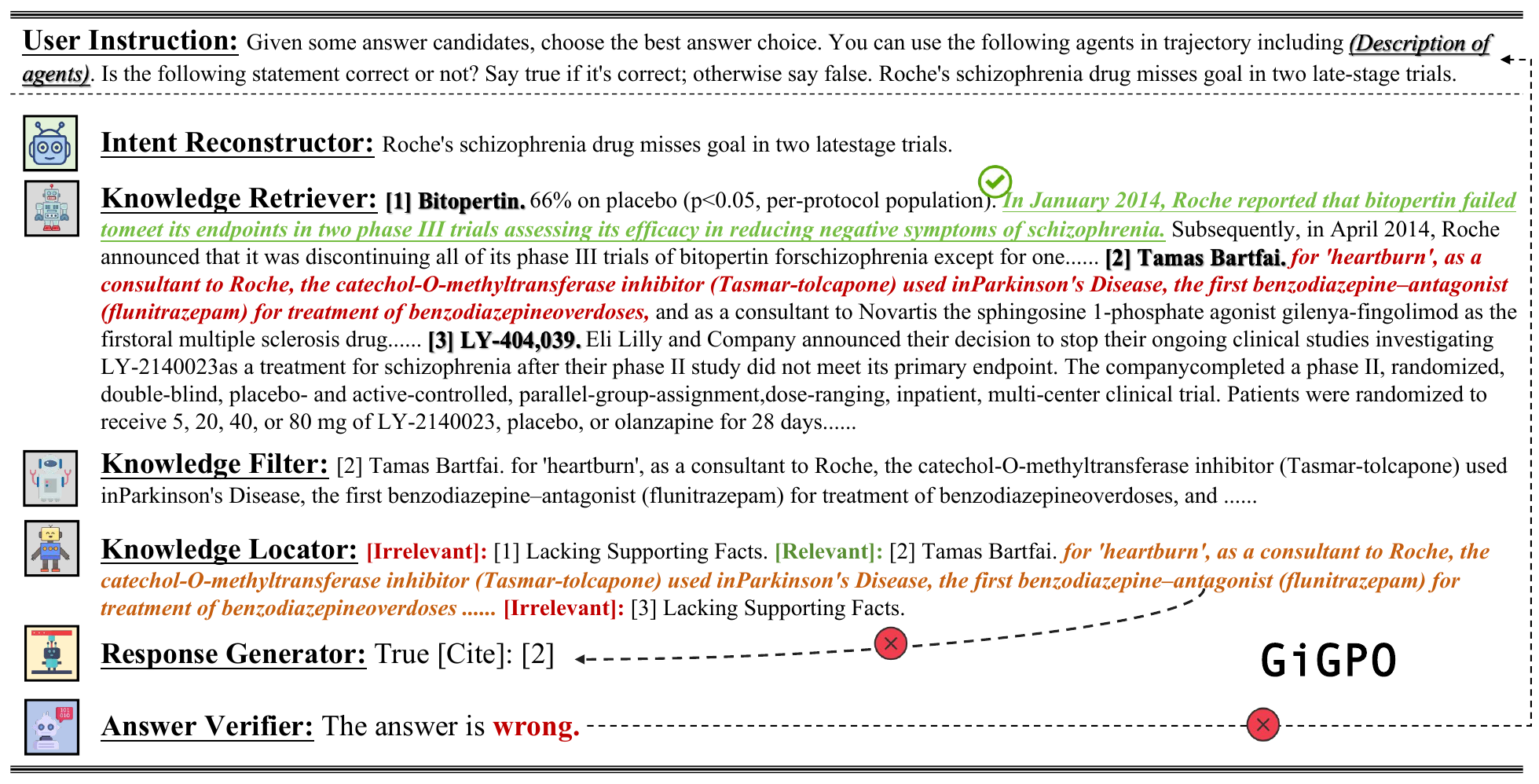}
  \caption{Complete response example from GiGPO.}
  \label{case_study3}
\end{figure*}

\subsection{LLMs' Backbone Analysis}
\label{backbone_result}

Our \emph{AMATA} framework decouples the model architecture from backbone selection.
We further evaluate several state-of-the-art LLMs, including Llama-3~\cite{DBLP:journals/corr/abs-2407-21783} and GPT-4o~\cite{gpt4_}, to validate the effectiveness of our method, using Llama-3-Instruct (8B) as the backbone.
As shown in Table~\ref{backbone_experiments}, trajectory planning methods based on RL demonstrate even greater potential when enhanced LLM capabilities are available.
We speculate that stronger foundational abilities are unlocked by post-RL training, which intrinsically provides planning skills for multi-agent tasks~\cite{DBLP:journals/corr/abs-2405-11106}.
Our \emph{AMATA} consistently exhibits performance improvements as the underlying backbone is strengthened.

\subsection{Case Study}
In this section, we present a case study analyzing our \emph{AMATA} framework alongside other trajectory training paradigms, including \emph{MMAgent}, \emph{SMART}, and \emph{GiGPO}.

The complete response generated by \emph{AMATA} is shown in Figure~\ref{complete_example}.
The user instruction pertains to retrieved document index ``[1]'', enabling $\mathcal{A}_{\text{KF}}$ and $\mathcal{A}_{\text{KL}}$ to make accurate inferences based on the document.
Moreover, sufficient external documentation supports the LLM's inference, resulting in high confidence in the generated answer.
Our answer verifier, $\mathcal{A}_{\text{AV}}$, confirms that no further validation is required.

In contrast, other trajectory learning paradigms face various types of errors.
As shown in Figure~\ref{case_study1}, since \emph{MMAgent} trains each agent independently, internal connections within the trajectory are neglected.
This causes the retriever to fail to fully comprehend the semantics of the user's question, especially when retrieved documents are labeled ``None''.
Consequently, the combined operation of knowledge retriever $\mathcal{A}_{\text{KR}}$, $\mathcal{A}_{\text{KF}}$, and $\mathcal{A}_{\text{KL}}$ does not effectively contribute to the LLM's inference.

As shown in Figure~\ref{case_study2}, SMART benefits from predefined global-local trajectory training and produces a generally correct trajectory path.
However, both SMART and GiGPO, as shown in Figures~\ref{case_study2} and \ref{case_study3}, fail to account for dependencies between inter-agent processes.
This oversight introduces external noise into the retrieved documents, resulting in inaccurate overall trajectories.

\section{Inference}
\label{inference_process}
Algorithm~\ref{algo} gives an overview of inference in our \emph{AMATA} framework.
During inference, \emph{AMATA} first analyzes the query to determine whether external knowledge is required.
If not, it directly generates and verifies the answer.
Otherwise, it retrieves and filters relevant documents, then generates a grounded response.
The answer is subsequently verified; if found incorrect, the process iterates with updated instructions to refine the response.
The specific steps are as follows:
\begin{itemize}
    \item \textbf{Step 1:} We utilize the intent reconstructor $\mathcal{A}_{\text{IR}}$ to decompose the user prompt $\mathcal{Q}$. If the initial trajectory head token $h = h_{\text{RG}}$ indicates that the question can be answered directly by the LLM without external knowledge, the response agent $\mathcal{A}_{\text{RG}}$ generates the answer $\mathcal{Y}$. Subsequently, the answer verifier checks the consistency of the generated response. \textbf{(Lines 1--13)}
    \item \textbf{Step 2:} When $h = h_{\text{KR}}$ indicates that the question $\mathcal{Q}$ requires external knowledge, the knowledge retriever agent $\mathcal{A}_{\text{KR}}$ retrieves external documents $\{d_1, \dots, d_{k \cdot m}\}$. The knowledge filter $\mathcal{A}_{\text{KF}}$ removes noise from these documents to produce a refined set $D = \{d_1, \dots, d_{w}\}$. The knowledge locator $\mathcal{A}_{\text{KL}}$ extracts fine-grained text spans $y_{\text{KL}}$ from $D$ based on $\mathcal{Q}$. \textbf{(Lines 16--23)}
    \item \textbf{Step 3:} If the extracted span is relevant ($r = \text{[Relevant]}$), the response generator $\mathcal{A}_{\text{RG}}$ formulates the answer using this span; otherwise, it answers directly. The answer verifier evaluates the response $\mathcal{Y}$ for factual correctness. If incorrect, \emph{AMATA} restarts the trajectory generation at \textbf{``Start''}, incorporating the erroneous response into $\mathcal{Q}$ for further reasoning. \textbf{(Lines 24--35)}
\end{itemize}

When the verifier determines that the LLM response is incorrect (i.e., \textbf{Lines 12} and \textbf{34}), we incorporate the incorrect answer into the prompt $\mathcal{Q}$ and reuse the previously generated dynamic trajectory for further inference, denoted as ``\textbf{goto Start}.''
To prevent repeated incorrect answer generation and infinite loops, a maximum of three iterations is allowed, after which the result is returned as the default answer.

Additionally, we analyze the out-of-domain generalization of our algorithm.
Once trained, \emph{AMATA} operates as a zero-shot, self-adaptive system that requires no pre-existing training traces or score annotations for new domains or questions.

\vspace{0.4em}
\noindent \textbf{Generalization to Unseen Datasets and Domains.} The core of \emph{AMATA}'s generalization lies in its learning objective:
We train it not to memorize specific answers or trajectories but to learn two fundamental principles:
\begin{itemize}
    \item \textbf{Intra-Trajectory Preference:} Dynamically assess which agents are important for a given question based on semantic content.
    \item \textbf{Inter-Agent Dependency:} Orchestrate selected agents in a coherent and efficient sequence.
\end{itemize}
Once this ``collaboration policy'' is learned, it can be applied to new questions. The model analyzes each new, unseen question at inference and uses its acquired knowledge to:
\begin{itemize}
    \item Dynamically predict the relevance of each agent (simulating internal scoring).
    \item Execute an optimal trajectory of agent invocations based on these predictions and learned dependencies.
\end{itemize}

\noindent \textbf{Applicability without Training Traces.} Our model incurs a one-time cost for bootstrapping collaborative reasoning capability. The trained framework is fully self-sufficient at inference.

\begin{algorithm}[!tb]
    \newcommand{\comm}[1]{\textcolor{gray!50}{\textit{#1}}}
    \caption{Inference of \emph{AMATA}}
    \label{alg_inference}
    \textbf{Require:} Intent Reconstructor $\mathcal{A}_\text{IR}$, Knowledge Retriever $\mathcal{A}_\text{KR}$, Knowledge Filter $\mathcal{A}_\text{KF}$, Knowledge Locator $\mathcal{A}_\text{KL}$, Response Generator $\mathcal{A}_\text{RG}$, Answer Verifier $\mathcal{A}_\text{AV}$, passage collections $d_1, \dots, d_k$, trajectory head token $h$, trajectory end token $e$ \\
    \textbf{Input}: User prompt $\mathcal{Q}$ \\
    \textbf{Output}: Answer $\mathcal{Y}$ \\ \rule{\linewidth}{0.1pt}
    \begin{algorithmic}[1]
        \State \textbf{Start:} $\mathcal{A}_\text{IR}$ predicts $q_1, \dots, q_m, e_{\text{IR}}, h$ given $\mathcal{Q}, h_{\text{IR}}$
        % \State \comm{\# \texttt{AMATA} can answer the question $\mathcal{Q}$ directly.}
        \If{$h = h_{\texttt{RG}}$}
            \State $\mathcal{A}_\text{RG}$ predicts $\mathcal{Y}$, $e_{\text{RG}}$, $h_{\text{AV}}$ given $\mathcal{Q}$, $e_{\text{IR}}$, $h_{\text{RG}}$
            \State \comm{\# \texttt{$\mathcal{A}_\text{AV}$ verifies the response.}}
            \State $\mathcal{A}_\text{AV}$ predicts $y_{\text{AV}}$, $e_{\text{AV}}$ given $\mathcal{Q}$, $\mathcal{Y}$, $h_{\text{AV}}$
            \If{$y_{\text{AV}} = \text{``Correct''}$}
                % \State \comm{\# \texttt{The trajectory is completed.}}
                \State \Return $\mathcal{Y}$
            \Else
                \State \comm{\# \texttt{Re-executing with wrong \text{Ans} $\mathcal{Y}$}.}
                \State \textbf{goto Start}
            \EndIf
\Else
        %     \State $\mathcal{A}_\text{G}$ predicts $y$, $e_g$ given $x$, $h_g$
        % \EndIf
        \State \comm{\# \texttt{$h=h_{\text{KR}}$ answer the question $\mathcal{Q}$.}}
        \For{each $p$ in $q_1, \dots, q_m$}
            \State Retrieve $(d_1, \dots, d_k)$ using $\mathcal{A}_\text{KR}$ given $p$, top-$k$
        \EndFor
        \State $D = \{d_1, \dots, d_{k \cdot m}\}$
        \State \comm{\# \texttt{$\mathcal{A}_{\text{KF}}$ filters out unrelated $d_i$}.}
        \State Filter $D = \{d_1, \dots, d_{w}\}$ using $\mathcal{A}_{\text{KF}}$ given $\mathcal{Q}$, $q$
        \State \comm{\# \texttt{$\mathcal{A}_{\text{KL}}$ locates the key text span $y_\text{KL}^{i}$.}}
        \State $\mathcal{A}_\text{KL}$ predicts $ \{(r_1, y_\text{KL}^{1}), \dots, (r_w, y_\text{KL}^{w})\}, e_{\text{KL}}, h_\text{RG}$ given $\mathcal{Q}$, $\{d_1, \dots, d_w\}$, $e_{\text{KR}}$, $h_{\text{KL}}$
        \If{$r = \text{[Relevant]}$}
            \State $\mathcal{A}_\text{RG}$ predicts $\mathcal{Y}$, $e_{\text{RG}}$, $h_{\text{AV}}$ given  $\mathcal{Q}$,$e_{\text{KL}}$, $h_{RG}$, $\{(r_1, y_\text{KL}^{1}), \dots, (r_w, y_\text{KL}^{w})\}$
        \Else
            \State $\mathcal{A}_\text{RG}$ predicts $\mathcal{Y}$, $e_{\text{RG}}$, $h_{\text{AV}}$ given $\mathcal{Q}$, $h_\text{RG}$
        \EndIf
    \State \comm{\# \texttt{$\mathcal{A}_\text{AV}$ verifies the response.}}
    \State $\mathcal{A}_\text{AV}$ predicts $y_{\text{AV}}$, $e_{\text{AV}}$ given $\mathcal{Q}$, $\mathcal{Y}$, $h_{\text{AV}}$
    \If{$y_{\text{AV}} = \text{``Correct''}$}
        \State \Return $\mathcal{Y}$
    \Else
        \State \textbf{goto Start}
    \EndIf
\EndIf
    \end{algorithmic}
    \label{algo}
\end{algorithm}

\end{document}